\documentclass[sigconf, nonacm]{acmart}

\newcommand\vldbdoi{XX.XX/XXX.XX}

\newcommand\vldbavailabilityurl{URL_TO_YOUR_ARTIFACTS}
 
\usepackage{amsmath,amsfonts,bbm}
\usepackage{algorithmic}
\usepackage{graphicx}
\usepackage{textcomp}
\usepackage{xcolor}

\usepackage{subfigure}
\usepackage{caption}
\usepackage{rotating}
\usepackage{algorithm}
\usepackage{booktabs}
\usepackage{multirow}
\usepackage{lipsum}
\usepackage{amsmath}
\usepackage{mathrsfs}
\usepackage{makecell}
\usepackage{float}

\newtheorem{definition}{Definition}[section]

\def\BibTeX{{\rm B\kern-.05em{\sc i\kern-.025em b}\kern-.08em
    T\kern-.1667em\lower.7ex\hbox{E}\kern-.125emX}}
\begin{document}

\title{A principled distributional approach\\ to trajectory similarity measurement
}

\author{Yufan Wang, Kai Ming Ting, Yuanyi Shang}
\affiliation{%
  \institution{Nanjing University}
  \city{Nanjing}
  \country{China}
}
\email{{wangyf,tingkm,shangyy}@lamda.nju.edu.cn}

\begin{abstract}
Existing measures and representations for trajectories have two longstanding fundamental shortcomings, i.e., they are computationally expensive and they can not guarantee the `uniqueness' property of a distance function: $dist(X,Y) = 0$ if and only if  $X=Y$, where $X$ and $Y$ are two trajectories.
This paper proposes a simple yet powerful way to represent trajectories and measure the similarity between two trajectories using a distributional kernel to address these shortcomings. It 
is a principled approach based on kernel mean embedding which has a strong theoretical underpinning. It has three distinctive features
in comparison with existing approaches. (1)
A distributional kernel is used for the very first time for trajectory representation and
similarity measurement. (2) It
does not rely on point-to-point distances which are used in most existing distances for trajectories. (3) It
requires no learning, unlike existing learning and deep learning approaches. We show the generality of this new approach in three applications: (a) trajectory anomaly detection, (b) anomalous sub-trajectory detection, and (c) trajectory pattern mining. We identify that the distributional kernel has (i) a unique data-dependent property and the above uniqueness property which are the key factors that lead to its superior task-specific performance; and (ii) runtime orders of magnitude faster than existing distance measures.
\end{abstract}


\maketitle


\ifdefempty{\vldbavailabilityurl}{}{
\vspace{.3cm}
\begingroup\small\noindent\raggedright\textbf{Artifact Availability:}\\
The source code, data, and/or other artifacts are available at \url{https://github.com/IsolationKernel/Codes/tree/main/IDK/TrajectoryDataMining}.
\endgroup
}

\section{Introduction}


With the rapid development of location technology, a vast amount of trajectory data has been produced, leading to growing interest in trajectory mining. Potential benefits of trajectory mining are in urban planning, transportation management and public safety.



Like other data mining tasks, similarity measurements for trajectories are a core operation in trajectory data mining. Commonly used measures include Dynamic Time Warping (DTW) \cite{DTW-FirstPaper-1971}, Hausdorff distance \cite{Hausdorff-Dist-Variational-Analysis-Book}, Fr\'{e}chet distance \cite{Frechet-Distance-1994} and edit distances \cite{ERP-EditDist-VLDB-2004,EDR-EditDist-SIGMOD2005}. They are all based on point-to-point distances between two trajectories, despite having different ways to accumulate multiple point-to-point distances between two trajectories to produce the final distance. Most importantly, these traditional measures ignore the distribution of points in a trajectory, which can lead to an unreasonable result in some cases (see Section \ref{sec_data_dependent_property} for details).


All the above measures have two longstanding fundamental shortcomings. First, they are computationally expensive, i.e., they have high time complexity of $O(m^2)$ in making a measurement between two trajectories, each having $m$ data points.
Second, none of the measures can guarantee the `uniqueness' property of a distance function: $dist(X,Y) = 0$ if and only if  $X=Y$, where $X$ and $Y$ are two trajectories 
(see e.g., \cite{AnomalousTD-ACCV2018,t2vec-ICDE2018}). As a result, they may produce some `irregularities' such as two different trajectories having a small distance or two similar trajectories having a large distance. Despite recent efforts that produce learned measure/representation \cite{AnomalousTD-ACCV2018,t2vec-ICDE2018}, they still could not guarantee to have gotten rid of this kind of `irregularities'. Even deep learning methods (e.g., \cite{GM-VSAE-ICDE2020,EncDec-AD,AnomalyTransformer}) do not provide any improvement on this front.

We are motivated to address these two shortcomings using a principled approach, which has three distinctive features in comparison with the above-mentioned approaches. First, a distributional kernel is used for the very first time to represent trajectories and measure the similarity between two trajectories. Second, 
it does not rely on point-to-point distances. Third,  it requires no learning. Yet, it performs as well as or better than existing measures and deep learning methods. 

Our contributions are:
\renewcommand{\theenumi}{\arabic{enumi}}
\begin{enumerate}
    \item Introducing a distributional kernel for trajectory representation and similarity measurement which has (i) linear time complexity $O(m)$ for measuring two trajectories which have a maximum length of $m$; and (ii) a strong theoretical underpinning based on kernel mean embedding \cite{HilbertSpaceEmbedding2007, KernelMeanEmbedding2017}.
    \item Analyzing the essential properties of a good measure for trajectories and identifying the importance of the data-dependent property of a measure for applications such as anomalous trajectory detection.
    

    
    \item Proposing simple and effective algorithms in three applications: anomalous trajectory \& sub-trajectory detections and frequent trajectory pattern mining, based on the powerful distributional kernel.
    
    
    \item Showing for the first time that the distributional measure can be applied successfully to four existing point anomaly detectors to detect anomalous trajectories.

    \item Conducting three empirical evaluations in three different applications to assess the effectiveness and efficiency of (a) the proposed distributional kernel, three commonly used distance measures and one representation learning, and (b) the proposed algorithms in comparison with existing algorithms in these applications.
\end{enumerate}

Our approach is distinguished from the other measures mentioned above in three key aspects:
\renewcommand{\theenumi}{\Roman{enumi}}
\begin{enumerate}
    \item A distributional kernel is used to represent each trajectory without learning and measuring the similarity between two trajectories.
    In contrast, existing works focus on some point-to-point distance based measures or learned representations. 
    \item The proposed distributional approach inherits the theoretical fundamentals of kernel mean embedding \cite{KernelMeanEmbedding2017}. Specifically, the resultant representation $\Phi: \mathbb{P} \rightarrow \mathscr{H}$ (Hilbert space)  is injective, i.e., $\parallel \Phi(\mathcal{P}_X) - \Phi(\mathcal{P}_Y) \parallel_{\mathscr{H}} = 0$ iff $\mathcal{P}_X = \mathcal{P}_Y$, where $\mathcal{P}_X, \mathcal{P}_Y \in \mathbb{P}$ and $\mathbb{P}$ is a set of probability distributions on $\mathbb{R}^\mathbf{d}$ \cite{Injective-2004}. Note that this is equivalent to the uniqueness property of a measure (without mapping $\Phi$) mentioned earlier. None of the existing measures and representation learning methods have been shown to have this property (see the next section for details).

  \item An implementation of the approach called Isolation Distributional Kernel ($\mathcal{K}_I$)  has a unique data-dependent property: \textbf{two distributions are more similar to each other when measured by $\mathcal{K}_I$ derived from a sparse region than that from a dense region}
  \cite{ting2020-IDK}. It enables $\mathcal{K}_I$-based detector to  gain higher detection accuracy than existing deep learning methods. Furthermore, this implementation is more efficient than traditional distance-based methods.
 \end{enumerate}





The rest of the paper is organized as follows. 
Section \ref{sec_related_work} reviews existing similarity/distance measures for trajectories. Section \ref{sec_current-understanding} presents the current work on distributional kernels and how they are used for point and group anomaly detections. 
Section \ref{sec_definitions} and \ref{sec_properties} describes (a) the intuitive assumptions that make distributional kernels a suitable candidate for measuring similarity between trajectories, 
and (b) the necessary properties of a similarity measure for trajectories.
The proposed distributional kernel-based algorithms for three applications are provided in Section \ref{sec_proposed_algorithms}. The empirical settings and evaluations are reported in Sections \ref{sec_settings} \& \ref{sec_experiments}, followed by the discussion and conclusion sections.

\section{Related Work}
\label{sec_related_work}
Existing trajectory similarity/distance measures can be divided into four categories: traditional measures, tailored measures and representation learning, and task-specific end-to-end approach. Examples of conventional measures include Dynamic Time Warping (DTW) \cite{DTW-FirstPaper-1971}, Hausdorff distance \cite{Hausdorff-Dist-Variational-Analysis-Book}, Fr\'{e}chet distance \cite{Frechet-Distance-1994} and edit distances \cite{ERP-EditDist-VLDB-2004,EDR-EditDist-SIGMOD2005}, longest common subsequence \cite{lcss2002}. Many of these measures compute their final distances between two trajectories based on some best-match point-pairs from the two trajectories. Dynamic programming is often used to find the matched pairs based on some criterion to determine the goodness of a match. 
High time complexity is the key limitation of these measures.

Note that some of these measures are set-based measures, e.g., Hausdorff and Fr\`{e}chet distances, where time/order information is ignored.

The second category is tailored measures which are motivated from the fact that existing distance measures such as DTW, Hausdorff and Fr\`{e}chet distances have some irregularities in measuring the distance between trajectories (see e.g., \cite{AnomalousTD-ACCV2018,t2vec-ICDE2018}).

Constructing a tailored measure often involves a process to learn the order/time dependent structure in a dataset of trajectories. 
An example method enlists RNN Autoencoder to learn a tailored measure \cite{AnomalousTD-ACCV2018} by minimizing the reconstruction errors between the input sequences and the constructed sequences.
Another recent deep learning work ST2Vec \cite{ST2Vec} aims to speed up the computation of an existing distance measure such as Hausdorff and Fr\`{e}chet distances by learning an approximate measure.  

The third category is representation learning which aims to learn a vector representation of trajectories via some transformation. Examples are  deep representation learning \cite{t2vec-ICDE2018}, the tube-droplet method \cite{Tube-Droplet-PAMI2017}, time-sensitive Dirichlet process mixture model \cite{tDPMM-PMAI2013} and hidden Markov model \cite{Traj-learning-PAMI2011}. 
We refer the readers to a recent survey of the representation and measures used for trajectories \cite{Traj_similarity-survey-2020} for further details.

It is interesting to note that none of these existing measures and learned methods have been shown to have the uniqueness property, i.e.,  $dist(X,Y) = 0$ if and only if  $X=Y$, as we mentioned in the last section. We postulate that the irregularities identified are largely due to the lack of this property.

To deal with a specific data mining task with trajectories, any existing point-based methods, which employ a distance/kernel function, can be used directly with a minimal change.  
For example, to deal with trajectory anomaly detection,  existing anomaly detectors such as LOF\cite{LOF-2000} and OCSVM\cite{OCSVM2001} can be employed directly to detect anomalous trajectory by simply replacing the distance/kernel function used with any of the above-mentioned measures. 

The fourth category is the task-specific end-to-end approach. While it is end-to-end, representation learning is one of its key learning problems. For example in end-to-end  anomaly detection, ATD-RNN \cite{ATD-RNN} is a supervised method based on RNN. IGMM-GAN \cite{IGMM-GAN} combines a Gaussian mixture model with GAN. Through
estimation of a generative probability density on the space of
human trajectories, IGMM-GAN generates realistic synthetic datasets and simultaneously facilitates multimodal anomaly detection. The semi-supervised  GM-VSAE \cite{GM-VSAE-ICDE2020} first converts each trajectory to a series of tokens, 
and then employs RNN. EncDec-AD\cite{EncDec-AD} combines LSTM with an autoencoder. It uses the reconstruction error as the anomaly score, assuming that normal data is easier to reconstruct than anomalous data. All these methods use (different kinds of) deep learning as the main tool for representation learning.


In summary, traditional distance measures have high computational cost and some irregularities which affect the effectiveness of the measurements. While some learning methods, which attempt to tradeoff effectiveness with efficiency, resolve the runtime issue only, they actually make the effectiveness issue worse because only an approximation of the intended measure is learned. 

Most importantly, existing methods of all four categories have not been shown to have the uniqueness property  which is necessary in ensuring  good performance in a specific task. 

It is interesting to note that the distributional information freely available in trajectories have been ignored in existing methods, so as the data dependent property of a measure (see Section \ref{sec_data_dependent_property} later). We find that both are powerful information and property for similarity measurement as well as for dealing with a specific task in order to gain good task-specific performance.



\vspace{6mm}
In the next section, we provide a brief description of the distributional kernel and how it is used to detect point anomalies and group anomalies in the current literature.

Table \ref{tab:symbols} shows the key notations used in this paper.
\begin{table}[h]
		\centering
		 \tabcolsep=2mm
		\begin{tabular}{ll}
	\hline
    $x$ & A point in $\mathbf{d}$-dimensional real domain $\mathbb{R}^\mathbf{d}$\\
    $\kappa$  & Isolation/Gaussian kernel\\
    $\phi$ & Kernel map of $\kappa$\\
    $X$ &  A trajectory of $\ulcorner x_1,\dots,x_\mu \urcorner$ with $|X|=\mu$ points\\
	$\mathcal{P}_X$ & Probability distribution that generates $x \sim \mathcal{P}_X$\\
	${\mathcal{K}}_I$ or ${\mathcal{K}}_G$ & Isolation/Gaussian Distributional Kernel\\
    ${\Phi}$ & Kernel mean map of ${\mathcal{K}}_I$ or ${\mathcal{K}}_G$\\
	$\mathbf{g}$ & Mapped point $\mathbf{g}={\Phi}(\mathcal{P}_X)$ in Hilbert space $\mathscr{H}$\\
	
	$D$ & Set of $n$ trajectories $\{X_i,  i=1,\dots,n\}$\\
	$\Pi$ &  Set of mapped points $\{ \mathbf{g}_i, i=1,\dots,n\}$ in $\mathscr{H}$\\
	F$_I$ or F$_G$ & Detector F employing $\Phi$ derived from ${\mathcal{K}}_I$ or ${\mathcal{K}}_G$\\
    $d_{W}$, $d_{H}$, $d_{F}$ & DTW, Hausdorff, Fr\'{e}chet distances \\
    \hline
	\end{tabular}
	\caption{Key symbols and notations used}
	\label{tab:symbols}
\end{table}

\section{Current understanding of distributional kernel and its applications in anomaly detection}
\label{sec_current-understanding}
\subsection{Distributional kernel}
\label{sec_distributional_kernel}
A distributional kernel measures the similarity between two distributions.
Let $X$ and $Y$ be two sets of iid samples generated from two distributions $\mathcal{P}_X$ and $\mathcal{P}_Y$, respectively. Based on kernel mean embedding (KME) \cite{KernelMeanEmbedding2017}, a distributional kernel is defined as follows:
\begin{equation}
   \mathcal{K}_G(\mathcal{P}_X,\mathcal{P}_Y) = \frac{1}{|X||Y|}\sum\limits_{x \in X, y \in Y} \kappa(x,y)
\label{eq:G_KME}
\end{equation}

While $\kappa$ is typically a Gaussian kernel in the KME framework \cite{KernelMeanEmbedding2017}, a recent work \cite{ting2020-IDK} has shown that using Isolation Kernel (IK) \cite{ting2018IsolationKernel} is a better option for point anomaly detection. 

A new distributional kernel \cite{ting2020-IDK} constructed by replacing the Gaussian kernel with IK in KME is briefly described as follows. 

Given a dataset $\mathsf{D} \subset \mathbb{R}^\mathbf{d}$,
IK derives a finite-dimensional feature map $\phi$ from $\mathsf{D}$, i.e., $\kappa(x,y|\mathsf{D}) = \left<\phi(x|\mathsf{D}), \phi(y|\mathsf{D})\right>$.
Then, Eq (\ref{eq:G_KME}) can be re-expressed as:
\begin{equation}
\begin{aligned}
 \mathcal{K}_I(\mathcal{P}_X,\mathcal{P}_Y | \mathsf{D}) 
  = & \frac{1}{|X||Y|}\sum\limits_{x \in X, y \in Y} \kappa(x,y|\mathsf{D})\\
  = & \frac{1}{|X||Y|}\sum\limits_{x \in X, y \in Y} \left< \phi(x | \mathsf{D}), \phi(y | \mathsf{D}) \right> \\
  = &  \left< \Phi(\mathcal{P}_X | \mathsf{D}), \Phi(\mathcal{P}_Y | \mathsf{D}) \right>,\\
\end{aligned}
\label{eq:I_KME}
\end{equation}
and the kernel mean map $\Phi$ of $\mathcal{K}_I$, which maps a distribution estimated by a sample set $X$ in input space to a point in Hilbert space,  is given as:
\begin{equation}
\Phi(\mathcal{P}_X | \mathsf{D}) = \frac{1}{|X|} \sum\limits_{x \in X} \phi(x | \mathsf{D}).
\label{eq:KME}
\end{equation}




\subsection{Point \& group anomaly detectors based on $\mathcal{K}$}
\label{sec:kad}
\hspace{3mm}\textbf{Point Anomaly detection}: A point anomaly detector based on $\mathcal{K}_I$ called IDK$(x)$ for each point $x \in \mathsf{D}$ \cite{ting2020-IDK} is given as follows: 
\begin{equation}
\mbox{IDK}(x) = \mathcal{K}_I(\delta(x),\mathcal{P}_\mathsf{D} | \mathsf{D})
\label{eqn-IDK}
\end{equation}
where $\delta(x)$ is a Dirac measure which converts a point into a distribution.

IDK$(x)$ returns a similarity score of point $x$ with respect to the (unknown) distribution which generates the dataset $\mathsf{D}$. 

\textbf{Group Anomaly detection}: Given a dataset of groups of points, $\{H_1,\dots,H_m\}$ and $H_i \subset \mathbb{R}^\mathbf{d}$,  a group anomaly detector aims to identify the few groups which are different from the majority of the groups in dataset.

A group anomaly detector applies $\mathcal{K}_I$ in two levels \cite{ting2020-IDK-GroupAnomalyDetection}. The first level maps each group to a point in a Hilbert space, i.e., $\mathbf{g}=\Phi(\mathcal{P}_{H})$. Given the set of $m$ points $\Pi = \{ \mathbf{g}_1,\dots,\mathbf{g}_m \}$ in Hilbert space, IDK$(x)$ can be applied to detect the point anomalies, where each point anomaly in Hilbert space corresponds to a group anomaly in input space.  

Given that Gaussian kernel\footnote{Note that though Gaussian kernel has an infinite-dimensional feature map, one can use a kernel functional approximation method such as the Nystr\"{o}m method \cite{Nystrom_NIPS2000} to produce an approximate finite-dimensional feature map. Then, a similar dot product expression according to Eq (\ref{eq:KME}) can be produced.} can be used in place of Isolation Kernel at each of the two levels,
four variants of group anomaly detectors can be created. They are given in Table~\ref{tab:variants}. 

\begin{table}[h]
    \centering
    \begin{tabular}{cc}
    Level 1 mapping ($\Phi(\mathcal{P}_{H})$) & Level 2 detector\\ \hline
    $\mathbf{g}=\Phi_I(\mathcal{P}_{H})$  & IDK$_I(\mathbf{g})   = \mathcal{K}_I(\delta(\mathbf{g}),\mathcal{P}_{\Pi_\mathbf{g}} | {\Pi_\mathbf{g}})$  \\
    $\mathbf{g}=\Phi_I(\mathcal{P}_{H})$  & GDK$_I(\mathbf{g}) = \mathcal{K}_G(\delta(\mathbf{g}),\mathcal{P}_{\Pi_\mathbf{g}} | {\Pi_\mathbf{g}})$     \\
    $\mathbf{h}=\Phi_G(\mathcal{P}_{H})$  & IDK$_G(\mathbf{h})   = \mathcal{K}_I(\delta(\mathbf{h}),\mathcal{P}_{\Pi_\mathbf{h}} | \Pi_\mathbf{h})$  \\
    $\mathbf{h}=\Phi_G(\mathcal{P}_{H})$  & GDK$_G(\mathbf{h}) = \mathcal{K}_G(\delta(\mathbf{h}),\mathcal{P}_{\Pi_\mathbf{h}} | \Pi_\mathbf{h})$     \\
      \hline
    \end{tabular}
    \caption{Four variants of group anomaly detectors, where the subscripts $_I$ and $_G$ denote the use of distributional kernels based on Isolation kernel and Gaussian kernel, respectively;  $\Pi_\mathbf{g}$ and $\Pi_\mathbf{h}$ denote the sets of points $\mathbf{g}$ and $\mathbf{h}$, respectively.}
    \label{tab:variants}
\end{table}





In addition,
as $\mathcal{K}_I$ and $\mathcal{K}_G$ are generic kernels, they can also be combined with existing anomaly detectors such as LOF \cite{LOF-2000} and OCSVM \cite{OCSVM2001}, by simply replacing the Euclidean distance (used in k-nearest neighbor employed in LOF) and Gaussian kernel (used in OCSVM) with either $\mathcal{K}_I$ or $\mathcal{K}_G$, to enable them to detect group anomalies \cite{ting2020-IDK-GroupAnomalyDetection}.

In the next section, we show for the first time that $\mathcal{K}_I$ and $\mathcal{K}_G$ can be effectively used to represent trajectories (as groups of points) and measure similarity between two trajectories. 

Note that k-nearest neighbors used in LOF and effectively 1-nearest neighbor used in IDK (see Equation (\ref{eqn-IDK})) and GDK are ideal in examining the effectiveness of $\mathcal{K}_I$ and $\mathcal{K}_G$ in measuring the similarity of trajectories in anomalous trajectory detection task. Because of the use of the kernel, they become  k-most similar neighbors and 1-most similar neighbor, respectively.

We then examine the effectiveness of anomaly detectors IDK, GDK, LOF and OCSVM which employ $\mathcal{K}_I$ and $\mathcal{K}_G$ on anomalous trajectory detection in Section \ref{sec_experiments}.

\section{Intuition and Problem Formulation}
\label{sec_definitions}
In this section, we first state our intuition of treating each trajectory as a sample set generated from an unknown distribution, and then provide the problem formulation. 
\subsection{Assumptions and intuitive examples}
\label{sec_intuition}

\begin{figure}[htbp] 
       \subfigure[Trajectories in $\mathbb{R}^1$ \& time domain]{
       \includegraphics[width=.45\linewidth]{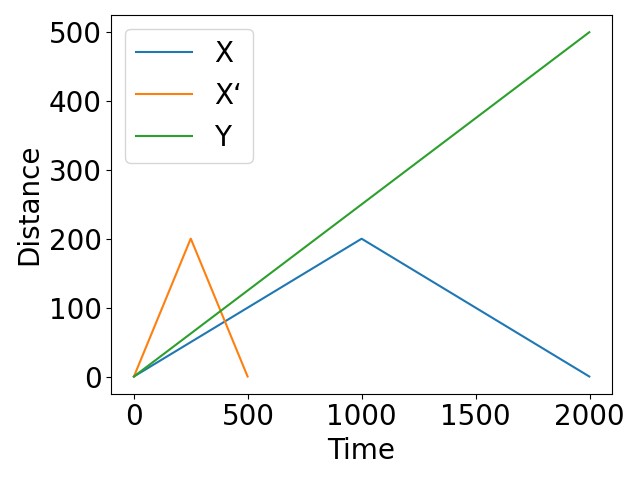}
       }
       \subfigure[pdfs of trajectories in $\mathbb{R}^1$ domain]{
          \includegraphics[width=.45\linewidth]{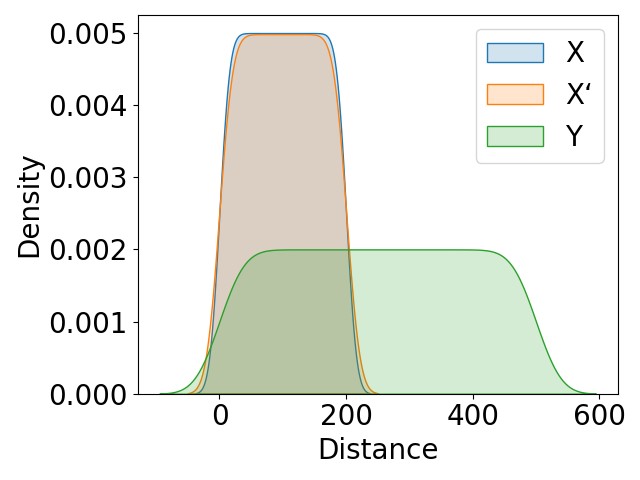}
       }
     \caption{Representing a set of trajectories in (a) 2-dimensional spatio-temporal space and (b) 1-dimensional spatial space in terms of pdfs. The pdfs are estimated using a kernel density estimator with the Gaussian kernel.}
    \label{fig:intuition_example}
\end{figure}

To use a distributional kernel to measure the similarity between two trajectories, we make the following assumptions:

\renewcommand{\theenumi}{\roman{enumi}}
\begin{enumerate}
  \item \textbf{Valid trajectories}: A valid trajectory  consists of a series of ordered points which obey the constraints in time and space.


  \item \textbf{Independent and identically distributed (iid) assumption}: A valid trajectory $X$ is assumed to be an iid sample set generated from an unknown probability distribution $\mathcal{P}_{X}$. 
  \item \textbf{Time is regarded to be one of the dimensions in $\mathbb{R}^\mathbf{d}$}:The time information (or order) of points in each trajectory can be included as a dimension, in addition to a spatial $\mathbf{d}-1$ dimensional space. 
  
\end{enumerate} 
With the above assumptions, all points in a valid trajectory $X$ can be seen as iid points $x \in \mathbb{R}^\mathbf{d}$ which are generated from an unknown probability distribution function (pdf) $\mathcal{P}_X$, i.e., $x \sim \mathcal{P}_X$, and
a distributional kernel can be used to compute the similarity between two trajectories effectively.
Here, we provide intuitive examples using a set of trajectories represented in $\mathbb{R}^2$ and $\mathbb{R}^1$ domains. 
\begin{itemize}
    \item $x \in \mathbb{R}^2$, where time is one of the two dimensions shown in Figure \ref{fig:intuition_example}(a): trajectory $X$ and $X'$ travel from the origin to a destination 200 meters away and then return to the origin at different constant speeds, and $Y$ travels from the origin to a destination 500 meters away. Here $X$, $X'$, and $Y$ have different distributions, i.e., $\mathcal{P}_X \ne \mathcal{P}_{X'} \ne \mathcal{P}_Y$. 
    \item $x \in \mathbb{R}^1$: When the time information is ignored, the estimations of the pdfs of $X$ and $X'$ are approximately the same, as shown in Figure~\ref{fig:intuition_example}(b). As a result, a distributional kernel $\mathcal{K}$ that measures the similarity between these two trajectories yields $\mathcal{K}(X,X') \approx 1$ because $\mathcal{P}_X \approx \mathcal{P}_{X'}$. In contrast, the similarity between $X$ and $Y$ yields $\mathcal{K}(X,Y) < \mathcal{K}(X,X')$ because $\mathcal{P}_X \ne \mathcal{P}_Y$.
\end{itemize}

Note that the above two representations can both be legitimate cases in practice. In applications where time is irrelevant in identifying a unique trajectory, it is unnecessary to include the time domain in the representation. 

\subsection{Problem formulations}
\begin{definition}
\textbf{Trajectory} $X$ is an ordered sequence of points, i.e. $X = \ulcorner x_1, \dots, x_i, \dots, x_\mu \urcorner $, where $i \in [1,\mu]$ indicates the order of traversal in $X$.  
\end{definition}
In practice, $x_i \in X$ is usually a GPS point having three attributes: longitude, latitude and timestamp.

\begin{definition}
\label{def-subt}
\textbf{Sub-trajectory} $X_s = \ulcorner x_{a},\dots,x_{b} \urcorner$ is a contiguous subsequence of $X$, denoted as $X_s \prec X$, where $1 \leq a \leq b \leq \mu$. 
\end{definition}

Note that $\ulcorner x_i \urcorner, i = 1, \dots, \mu $ are also sub-trajectories of $X$. We denote this special case of $\ulcorner x_i \urcorner $ as a \emph{point-sub-trajectory}.

\begin{definition}
\label{def-max-subt}
\textbf{Maximal sub-trajectory} $X_s \prec X$ is a contiguous subsequence in $X$ of maximal length if $x_{a-1}$ and $x_{b+1}$ cannot be valid members of  $X_s= \ulcorner x_{a},\dots,x_{b} \urcorner$ based on some criterion. 
\end{definition}

A trajectory may contain multiple maximal sub-trajectories separated by invalid members based on some criterion.

The problems of the three applications we considered in this paper are defined as follows:

\begin{definition}
\textbf{Anomalous trajectory detection}. Given a dataset $D=\{X_i, i=1,\dots,n\}$, anomalous trajectory detection aims to detect trajectories in $D$ which are rare and different from the majority. 
\end{definition}

\begin{definition}
\textbf{Anomalous sub-trajectory detection}. Given an anomalous trajectory $X$, anomalous sub-trajectory detection aims to detect all maximal sub-trajectories $X_s \prec X$ that make $X$ anomalous with respect to a given dataset of trajectories.
\end{definition}

\begin{definition}
\label{def-fp}
    \textbf{Frequent sub-trajectory pattern mining} aims to identify sub-trajectory patterns in which many sub-trajectories in a dataset of trajectories are in shared locations. The representatives of these shared maximal sub-trajectories are called sub-trajectory patterns.
\end{definition}


\section{Important Properties of Similarity Measures}
\label{sec_properties}

Table \ref{tab:property-comparison} presents four important properties of any measures for trajectories, i.e., uniqueness, point-to-point distance-based, distribution-based and data dependent. We describe the first three properties in the next subsection, and the details of the fourth property in the following subsection.

\begin{table}[h]
    \tabcolsep=2pt
    \centering
    \begin{tabular}{l|ccccc}
    \hline
     & $\mathcal{K}_I$ & $\mathcal{K}_G$ & $d_{W}$ & $d_H$ & $d_F$\\ \hline
    Uniqueness property & $\checkmark$ & $\checkmark$ & $\times$ & $\times$ & $\times$ \\
     Point-to-point distance-based &  $\times$ & $\checkmark$ & $\checkmark$ & $\checkmark$ & $\checkmark$\\
     Distribution-based & $\checkmark$ & $\checkmark$ & $\times$ & $\times$ & $\times$\\
     Data-dependent & $\checkmark$ & $\times$ & $\times$ & $\times$ & $\times$ \\ 
    \hline
     Time complexity & \multicolumn{2}{c}{ \scriptsize$m+n$} & \scriptsize$mn$ & \scriptsize$mn$ & \scriptsize$mn\log(mn)$\\
    \hline
    \end{tabular}
    \caption{Compliance with properties listed above of a trajectory similarity measure. Time complexity is for computing the similarity of two trajectories with size $m$ and $n$, respectively. $\mathcal{K}_I$ and $\mathcal{K}_G$ have the same time complexity $m+n$.  
    }
    \label{tab:property-comparison}
\end{table}

\subsection{Uniqueness, point-to-point distance-based and distribution-based properties}
The uniqueness property is essential for any distance function: $dist(X,Y) = 0$ if and only if  $X=Y$, where $X$ and $Y$ are two trajectories. An example with three trajectories is provided in Table \ref{tab:example_dtw_hd_fr} to examine whether the five measures comply with this property, where $X'$ is a translated version of $X$, $Y$ is 
 a different trajectory from either $X$ and $X'$.

\begin{table}[htbp]
   \begin{minipage}[t]{\linewidth}
    \centering
    \scriptsize
    \begin{tabular}{ll|l|l|c}
    \hline
    \multicolumn{2}{l|}{}        & ($X,X'$) & ($X,Y$) & $d(X,X')<d(X,Y)$\\
    \hline
     \multirow{3}{*}{I} & $d_{W}$ & \makecell[c]{.50} & \makecell[c]{.43} & $\times$\\
     & $d_{H}$ & \makecell[c]{.50} & \makecell[c]{.50} & $\times$\\
     & $d_{F}$ & \makecell[c]{.50} & \makecell[c]{.50} & $\times$\\
    \hline
    \multirow{2}{*}{II} & $\mathcal{K}_I$ & \makecell[c]{.48} & \makecell[c]{.23} & $\checkmark$\\
     & $\mathcal{K}_G$ & \makecell[c]{.45} & \makecell[c]{.27} & $\checkmark$\\
     \hline
    \end{tabular}
   \end{minipage}
    \begin{minipage}[t]{.4\linewidth}
    \centering
     \includegraphics[width=4cm]{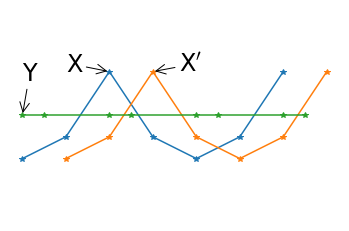}
    \end{minipage}
    \caption{An example of unreasonable results produced by the three distance measures $d_{W}$, $d_{H}$ \& $d_{F}$, in sharp contrast with distributional kernels $\mathcal{K}_I$ and $\mathcal{K}_G$. 
Row I presents distances, while row II presents similarities.
}
    \label{tab:example_dtw_hd_fr}
\end{table}

In this example, $d_{W}$, $d_{H}$ and $d_{F}$ produce distances for $X$ and $Y$ which are equal to or smaller than the distances for $X$ and $X'$. 

The reason for this unreasonable result is that: $d_{W}$, $d_{H}$, $d_{F}$ are based on point-to-point distances that  consider point-to-point matching only without considering the distribution of the points.

In contrast, $\mathcal{K}_I$ and $\mathcal{K}_G$ produce measurements which are consistent with the uniqueness property because they are both based on kernel mean embedding which have been shown theoretically to have this property (see \cite{KernelMeanEmbedding2017,ting2020-IDK} for the details).

\subsection{Data-dependent property of $\mathcal{K}_I$}
\label{sec_data_dependent_property}
Table \ref{tab:property-comparison} shows that only $\mathcal{K}_I$ has the data-dependent property among all the five measures. 
Given a trajectory dataset $D=\{X_i, i=1,\dots,n \}$, the data-dependent similarity measure $\mathcal{K}_I$ is derived from $\mathsf{D} = \cup_{i=1}^n X_i$ \cite{ting2020-IDK}. $\mathcal{K}_I$ relies on local neighborhoods which are large in the sparse region and small in the dense region. This leads directly to a unique data-dependent property \cite{ting2020-IDK}:
\begin{definition}
\textbf{Data-dependent property of $\mathcal{K}_I$}.
Two distributions are more similar to each other when measured by $\mathcal{K}_I$ derived from a sparse region than that from a dense region.
\end{definition}

This property is inherited from Isolation Kernel (IK) which has a similar data-dependent property \cite{ting2018IsolationKernel,IsolationKernel-AAAI2019}, as $\mathcal{K}_I$ \cite{ting2020-IDK} is built based on IK. The other four measures in Table \ref{tab:property-comparison} do not have the data dependent property because they all use a data independent measure such as Euclidean distance or Gaussian kernel.

Two scenarios, in which a data-dependent kernel such as IK has a significant impact on detection accuracy, are presented below.

\subsubsection{\textbf{Trajectories in dense and sparse clusters}}
\begin{figure}[h]
    \centering
    \includegraphics[width = .4\textwidth]{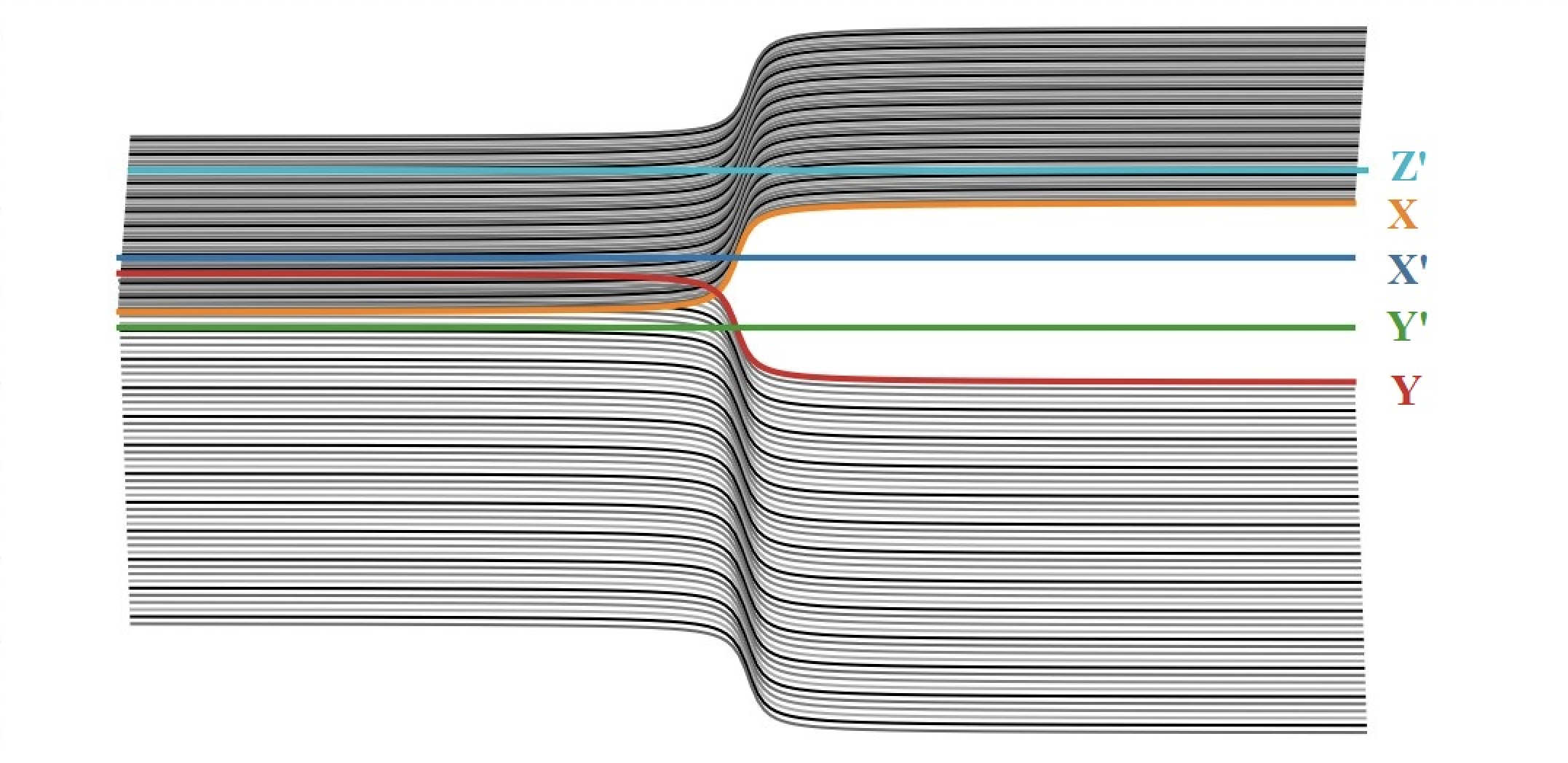}
    \caption{An example dataset. The trajectories are indexed, from \#0 to \#102 from top to bottom on the right. Three anomalous trajectories $Z'$, $X'$, and $Y'$, which have indices at \#40, \#51, \#52 respectively. On the right half of the trajectories, each pair of $X$ \& $X'$ and $Y$ \& $Y'$ has the same (vertical) Euclidean distance.
    }
    \label{fig:example-trajectories}
\end{figure}

The example dataset, shown in Figure \ref{fig:example-trajectories},
consists of 103 trajectories, with one normal dense cluster (top 50 trajectories), one normal sparse cluster (bottom 50 trajectories), and three anomalous trajectories $X'$, $Y'$, and $Z'$. Anomalous trajectories are all straight-line, whereas the normal trajectories in the two clusters are not straight-line trajectories. 

We apply anomaly detectors GDK$_G$ and IDK$_I$ (where they use $\mathcal{K}_G$ and $\mathcal{K}_I$, respectively; see Section \ref{sec:kad} for details) on the dataset shown in Figure~\ref{fig:example-trajectories} to illustrate the superior detection capability due to $\mathcal{K}_I$ than $\mathcal{K}_G$. 



\begin{figure}[h]
    \centering
    \includegraphics[width=.23\textwidth]{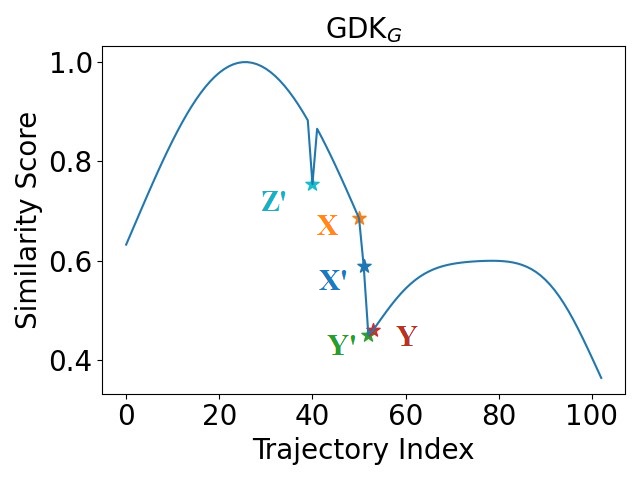}
    \includegraphics[width=.23\textwidth]{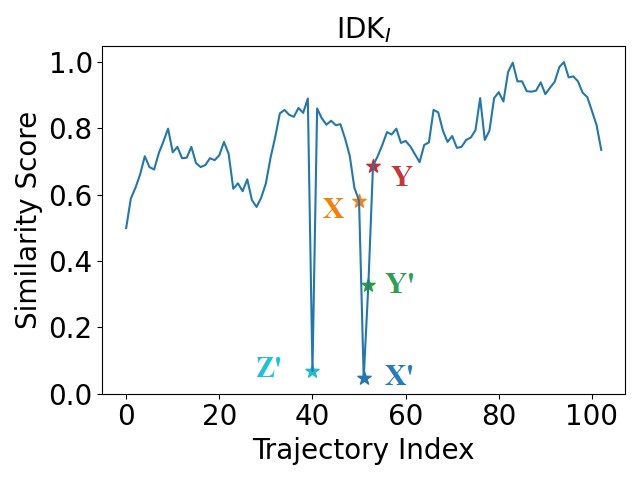}
    \caption{Similarity scores from GDK$_G$ and IDK$_I$ on trajectories shown in Figure~\ref{fig:example-trajectories}.}
    \label{fig:score-comparison}
\end{figure}

As illustrated in Figure \ref{fig:score-comparison}, GDK$_G$ exhibits a distribution of similarity scores which is reminiscent of a density distribution, where trajectory indices \#0-\#50 are members of the dense cluster, and indices \#53-\#102 are members of the sparse cluster. The three anomalous trajectories $Z'$, $X'$ and $Y'$ (indices \#40, \#51 \& \#52) have similarity scores close to the edges of the dense cluster. In addition, all trajectories in the sparse cluster have similarity scores less than the dense cluster, and the trajectory with the lowest similarity score is at index \#102 (i.e., the bottom trajectory in Figure~\ref{fig:example-trajectories}). As a result, almost all the normal trajectories in the sparse cluster have lower similarity scores than the two anomalous trajectories $Z'$, and $X'$. Thus, $Z'$ and $X'$ cannot be identified as anomalous trajectories by GDK$_G$.

In contrast, the distribution of the similarity score of IDK$_I$ is more balanced between the dense cluster and sparse cluster in Figure \ref{fig:score-comparison}, and all three anomalous trajectories have the lowest similarity scores. Thus, $Z'$, $Y'$, and $X'$ are easily identified as anomalous trajectories by IDK$_I$. 
\textbf{This scenario indicates that the data-dependent $\mathcal{K}_I$ is a better measure than the data-independent $\mathcal{K}_G$ for trajectory anomaly detection in datasets containing regions of varied densities}.

\subsubsection{\textbf{Sheepdogs trajectories}}
\begin{figure}[H]
    \centering
    \includegraphics[width=.38\textwidth]{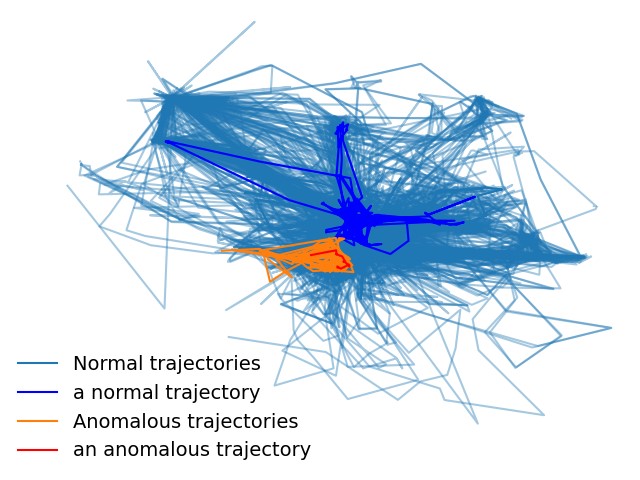}
    \caption{Sheepdogs trajectories in blue; sheep in orange.}
    \label{fig:sheepdogs}
\end{figure}

Here we use a real-world example, i.e., the \textit{Sheepdogs} dataset, which differs from the previous example in many aspects. The dense and sparse regions in \textit{Sheepdogs} have a significant impact on the detection accuracy of a detector that relies on a point-to-point distance measure.  Figure \ref{fig:sheepdogs} shows a total of 538 trajectories, of which 23 are trajectories of sheep and the rest are sheepdogs. 

The trajectories of sheep cluster in a small region; whereas each trajectory of sheepdogs is much longer and it travels around in a wide area. They are not `neat' artificial dense and sparse regions, as we have shown in Figure~\ref{fig:example-trajectories}. Each trajectory is hap-hazard and there is no clear grouping, especially with sheepdogs. This is a real-world example of a dense region of sheep trajectories lying within a wider region of sparse (and scattered) trajectories of sheepdogs. 

LOF (with three distance measures and $\mathcal{K}_G$) and GDK$_G$ (with $\mathcal{K}_G$) perform poorly on this dataset. They all have detection accuracy AUCs range between 0.56 and 0.93. In contrast, IDK$_I$ and LOF$_I$ which employ  $\mathcal{K}_I$ perform significantly better with AUC=0.98 and 0.99, respectively (see Table \ref{tab:exp_res}).

The \textit{Sheepdogs} dataset corresponds to the case of clustered anomalies found in point anomaly detection (see e.g., \cite{SCiForest-2010}), 
i.e., anomalous trajectories are clustered in a small region, while the majority of the (normal) trajectories are scattered in a wide area.

\section{Proposed Algorithms}
\label{sec_proposed_algorithms}
To show the generality of the proposed distributional kernel for trajectory representation and similarity measurement, we apply the distributional kernel $\mathcal{K}$ to three applications, i.e., trajectory anomaly detection, anomalous sub-trajectory detection and frequent trajectory pattern mining. The proposed $\mathcal{K}$ based algorithms in these applications are presented  in the next three subsections.



\subsection{Anomalous trajectory detection}
\label{sec_ano_trj_detection}
\begin{definition}
Given $D=\{X_i | i=1,\dots,n\}$, an anomalous trajectory $Q \in D$ is rare wrt $D$ and is generated from a pdf different from those generating the normal trajectories $X_i \in D$, that is, for most $i \in [1,n], \mathcal{P}_{X_i} \ne \mathcal{P}_{Q}$. 
\label{def:alg1}
\end{definition}

Following Definition \ref{def:alg1}, a distributional kernel $\mathcal{K}$ is used to represent each trajectory, and then to compute the similarity between two trajectories. 



In the light of Eq \ref{eq:KME}, the level-1 kernel mean map $\Phi(\mathcal{P}_{X}|D)$ from $\mathcal{K}$ that maps a trajectory $X$ to a point $\mathbf{g}$ in Hilbert space via the feature map $\phi$ built from $\mathsf{D} = \cup_{i=1}^n X_i$ is given as:
\begin{equation}
    \mathbf{g} = \Phi(\mathcal{P}_{X} | D) = \frac{1}{|X|} \sum_{x \in X} \phi(x|\mathsf{D}) 
\end{equation}

Let $\Pi= \{\mathbf{g}_1,\dots,\mathbf{g}_n\}$ be the set of $\Phi$-mapped points from $D$. 

With these representations, 
an existing point anomaly detector $\mathcal{F}(\cdot|\Pi)$ can be trained from $\Pi$, and then it provides the anomaly score for each $\mathbf{g} \in \Pi$, which corresponds to a trajectory in the given dataset $D$.

When IDK \cite{ting2020-IDK} is used as the detector $\mathcal{F}$, a score for any mapped point $\mathbf{g}$ with respect to $\Pi$  has the following expression (as used in Equations \ref{eq:I_KME}, \ref{eq:KME} \& \ref{eqn-IDK}):
\[
\mathcal{F}(\mathbf{g}|\Pi) = \mathcal{K}(\delta(\mathbf{g}), \mathcal{P}_\Pi) = \left< \Phi_2(\delta(\mathbf{g})), \Phi_2(\mathcal{P}_\Pi) \right>
\]
\noindent
where level-2 kernel mean map $\Phi_2(\mathcal{P}_\Pi) = \frac{1}{|\Pi|} \sum_{\mathbf{g} \in \Pi} \phi_2(\mathbf{g}|\Pi)$.

Note that $\Phi$ and $\Phi_2$ are derived from $D$ and $\Pi$, respectively. We drop `$|D$' and `$|\Pi$' for brevity hereafter.

\begin{algorithm}[h]
\caption{$\mathcal{K}$ for anomalous trajectory detection}
\label{alg:IDK2}
\hspace{-1.6cm}\textbf{Input}: Dataset of trajectories $D=\{X_i,\  i=1,\dots,n \}$; \\ 
\hspace{-0.2cm} kernel $\mathcal{K}(\cdot,\cdot) = \left< \Phi(\cdot), \Phi(\cdot) \right>$;
anomaly detector $\mathcal{F}$.\\
\hspace{-1.8cm}\textbf{Output}: List of $X_i,\  i=1,\dots,n$ ordered by score $\alpha_i$.\\
\begin{algorithmic}[1] 
\STATE \textcolor{blue}{* Map each trajectory $X_i \in D$ to a point in Hilbert space using the kernel mean map $\Phi(\mathcal{P}_{X_i})$} \\
For each $i=1,\dots,n,\ \mathbf{g}_i={\Phi}(P_{X_i})$
\STATE $\Pi = \{\mathbf{g}_i, i=1,\dots,n\}$
\STATE \textcolor{blue}{* Build a point anomaly detector $\mathcal{F}$ from $\Pi$ and get score $\alpha_i$ for each $\mathbf{g}_i$}\\
For each $i=1,\dots,n,\ \alpha_i=\mathcal{F}(\mathbf{g}_i|\Pi)$
\STATE Sort $X_{i} \in D$ in decreasing order by $\alpha_i$ if  $\alpha_i$ is an anomaly score (in ascending order if $\alpha_i$ is a similarity score)
\end{algorithmic}
\end{algorithm}

The anomalous trajectories in $D$ correspond to those points $\mathbf{g} \in \Pi$ which have the highest anomaly scores. The procedure described above is summarized in Algorithm \ref{alg:IDK2}. Note that this algorithm is a generalization of the IDK$^2$ \cite{ting2020-IDK-GroupAnomalyDetection} algorithm which admits any point anomaly detector to be used. Table \ref{tab:detectors} shows three examples that employ mapping function $\Phi$ derived from $\mathcal{K}_I$. Mapping function $\Phi$ derived from $\mathcal{K}_G$ can be similarly applied.

\begin{table}[h]
    \centering
    \tabcolsep=1mm
    \begin{tabular}{c|c}
    \hline
Detector $\mathcal{F}$ uses $\mathcal{K}_I$    &  Alg \ref{alg:IDK2}: line 4 ($\alpha_i$)  \\ \hline
  IDK$_I$ &  IDK$(\mathbf{g}_i|\Pi)$\\
  LOF$_I$ & LOF$(\mathbf{g}_i|\Pi)$ \\
  OCSVM$_I$ & OCSVM$(\mathbf{g}_i|\Pi)$\\
  \hline
    \end{tabular}
    \caption{Algorithm \ref{alg:IDK2} that incorporate existing point anomaly detectors IDK, LOF, and OCSVM, trained from $\Pi$.}
    \label{tab:detectors}
\end{table}

\subsection{Anomalous sub-trajectory detection}

A trajectory $Q$ is anomalous wrt a given set $D$ of normal trajectories if there are anomalous sub-trajectories in $Q$. We propose a simple yet effective algorithm based on $\mathcal{K}_I$ to detect the anomalous sub-trajectories that exist in an anomalous trajectory. The procedure is presented in Algorithm \ref{alg:subtraj-detector2}.  

The idea is to identify the individual point-sub-trajectories (recall Definition \ref{def-subt}) in the given anomalous trajectory $Q$ that make $Q$ anomalous with respect to $D$. Once the anomalous point-sub-trajectories are identified, the maximal sub-trajectories (Definition~\ref{def-max-subt}) are extracted from them to be the detected anomalous sub-trajectories $Q_s \prec Q$.

$\mathcal{K}_I$ is used to detect the anomalous point-sub-trajectories in $Q$ with respect to the average of kernel mean maps of all trajectories in  $D$. In the process, the kernel mean map $\Phi$ of $\mathcal{K}_I$ is used to map (a) each trajectory in $D$ in the input space into a point in Hilbert space; and (b) each point-sub-trajectories in $Q$ into a point in Hilbert space.
The same feature map $\Phi$, constructed based on the trajectories in $D$, performs both mappings.

\begin{algorithm}[h]
\caption{Anomalous sub-trajectory detection via map $\Phi$}
\label{alg:subtraj-detector2}
\hspace{-0.5cm} \textbf{Input}: Dataset of normal trajectories $D=\{X_i,\  i=1,\dots,n \}$; \\ 
\hspace{-1.1cm} kernel $\mathcal{K}(\cdot,\cdot) = \left< \Phi(\cdot), \Phi(\cdot) \right>$; threshold ${\tau}$;\\
\hspace{-1.2cm} anomalous trajectory $Q= \ulcorner  x_1,\dots,x_m  \urcorner $.\\
\hspace{-1cm} \textbf{Output}: All maximal anomalous sub-trajectories $Q_s \prec Q$.\\
\begin{algorithmic}[1] 
 \STATE \textcolor{blue}{* Map each trajectory $X \in D$ to a point in Hilbert space using the  kernel mean map $\Phi(\mathcal{P}_{X})$ derived from $D$} \\
 $\Pi = \{\mathbf{g}_i, i=1,\dots,n \}$, where $\mathbf{g}_i={\Phi}(\mathcal{P}_{X_i})$\\
 \STATE \textcolor{blue}{* Score each point-sub-trajectory $\ulcorner x \urcorner \prec Q$ wrt the average of kernel mean maps of all  $X \in D$.}\\
 For each $\ulcorner x \urcorner \prec Q, \beta_x= \left< \Phi(\delta(x)), \bar{\mathsf{g}}) \right>$, where $\bar{\mathsf{g}} = \frac{1}{n}\sum_{i=1}^{n}\mathsf{g}_i$  
\STATE $\mathsf{G} = \{ \ulcorner x \urcorner \prec Q \mid \beta_x  \leq {\tau} \}$
\STATE Extract every maximal sub-trajectory $Q_s \prec Q$ in $\mathsf{G}$.
\STATE Return all maximal anomalous sub-trajectories $\forall Q_s \prec Q$
\end{algorithmic}
\end{algorithm}

\subsection{Frequent sub-trajectory pattern mining}
Given a dataset $D$ of trajectories with a known number of clusters, frequent sub-trajectory pattern mining aims to extract sub-trajectory patterns that are frequented by many trajectories. 

Many existing methods are based on sub-trajectory clustering or sequence pattern mining. Here we show that this problem can be solved efficiently based on a distributional kernel.  The procedure is shown in Algorithm \ref{alg:fpm}, which has linear time complexity. 

The idea is similar to the task of anomalous sub-trajectory detection in two aspects: Each trajectory in the given dataset is mapped into a point in Hilbert Space; and a representative trajectory $X_r$ of each cluster (analogous to the given $Q$ in Algorithm \ref{alg:subtraj-detector2}) is used to discover the frequent sub-trajectory patterns in $X_r$.  

There are two additional works. First, the representative trajectory $X_r$ is to be discovered in each given cluster of trajectories. This is achieved by finding the trajectory in a cluster which is most similar to all trajectories in the cluster (see line 4 in Algorithm \ref{alg:fpm}). Second, the score of each point-sub-trajectory $\ulcorner x \urcorner \prec X_r$ is computed with respect to $\bar{\mathbf{c}}$, i.e., the average of kernel mean maps of all trajectories in $D$ (equivalent to all trajectories of all clusters). And we are interested in points in $X_r$ which are most similar to $\bar{\mathbf{c}}$ (the interest is in least similar points when detecting anomalous sub-trajectories in Algorithm~\ref{alg:subtraj-detector2}). This score is computed in line 5; and the most similar points are collected via a threshold $\gamma$ in line 6.  Line 7 in Algorithm~\ref{alg:fpm} simply extracts all the maximal sub-trajectory patterns $X_s \prec X_r$. They are patterns because they are extracted from the representative trajectory of a cluster, representing many sub-trajectories in $D$ (recall Definition \ref{def-fp}). 


\begin{algorithm}[h]
\caption{Frequent sub-trajectory pattern mining via map $\Phi$}
\label{alg:fpm}
\hspace{-0.9cm} \textbf{Input}: Dataset of normal trajectories $D=\cup_{i=1}^{k} C_i$, where $C_i = \{X_j, j=1, \dots, n_i\}$ is a cluster of trajectories; \\
\hspace{-1.2cm}  kernel $\mathcal{K}(\cdot,\cdot) = \left< \Phi(\cdot), \Phi(\cdot) \right>$; threshold $\gamma$. \\
\hspace{-0.9cm}\textbf{Output}: A set of all frequent sub-trajectory patterns $R$.\\
\begin{algorithmic}[1] 
\STATE $R = \emptyset$
\STATE \textcolor{blue}{* Map each $X \in C_i$ using the kernel mean map $\Phi(\mathcal{P}_{X})$, and compute the average of kernel mean maps of all $X \in C_i$} \\
For each $i=1,\dots, k,\ \mathbf{c}_i=\frac{1}{|C_i|}\sum_{X \in C_i} {\Phi}(\mathcal{P}_{X})$ \\
\FOR{$i = 1$ to $k$}
\STATE \textcolor{blue}{* 
Choose the most representative trajectory $X_r$ in $C_i$}\\
$X_r = \max_{X \in C_i} \left<\Phi(\mathcal{P}_{X}), \mathbf{c}_i\right>$\\
\STATE \textcolor{blue}{* Score each point-sub-trajectory $\ulcorner x \urcorner \prec X_r$ wrt the average of kernel mean maps of all $C_i, i=1,\dots,k$}\\
For each $\ulcorner x \urcorner \prec X_r, \theta_x = \left<\Phi({\delta(x)}), \bar{\mathbf{c}}\right>$, where $\bar{\mathbf{c}} = \frac{1}{k}\sum_{i=1}^{k}\mathbf{c}_i$\\
\STATE $\mathsf{G} = \{\ulcorner x \urcorner \prec X_r \mid \theta_x > \gamma \}$
\STATE Extract every maximal sub-trajectory pattern $X_s \prec X_r$ in $\mathsf{G}$ \\
\STATE $R = R \cup \{\forall {X_s \prec X_r}\}$
\ENDFOR
\STATE Return $R$ the set of frequent sub-trajectory patterns 
\end{algorithmic}
\end{algorithm}

\noindent
\textbf{Summary for Section \ref{sec_proposed_algorithms}}

The common ingredients in the above three algorithms are that (a) each trajectory $X$ or point-sub-trajectory $\ulcorner x \urcorner \prec Q$ in the input space is represented as a point in Hilbert space induced by distributional kernel $\mathcal{K}$ via its feature map $\Phi(\mathcal{P}_X)$ or $\Phi(\delta(x))$; (b) a group of trajectories is represented as an aggregate of their kernel mean  maps; and (c) the similarity between a trajectory/point-sub-trajectory and a group of trajectories is computed via a dot product of their mapped points in Hilbert space.

The key difference between Algorithm \ref{alg:IDK2} and Algorithms \ref{alg:subtraj-detector2} \& \ref{alg:fpm} is that the former deals with trajectories only and the latter intends to find maximal sub-trajectories of one or a few individual trajectories only. In Algorithm \ref{alg:IDK2}, level-2 kernel mean map $\Phi_2$ is required to represent the group of all (normal and anomalous) trajectories in $D$ in order to detect the anomalous trajectories. In Algorithms \ref{alg:subtraj-detector2} \& \ref{alg:fpm},  only an average of level-1 kernel mean maps of a group of trajectories is required, instead of level-2 kernel mean map. This is because the input to these algorithms are normal trajectories only (which can be identified by using Algorithm \ref{alg:IDK2}). 


\section{Experimental Design and Settings}
\label{sec_settings}
The experiments are designed to answer the following questions:
\begin{enumerate}
  \item Does $\mathcal{K}_I$ perform better than $\mathcal{K}_G$?  
  \item Is there any advantage of distributional kernel $\mathcal{K}$ over existing similarity measures and representation methods?
  \item Which is the best detector for anomalous trajectory detection and anomalous sub-trajectory detection? 
  \item Could $\mathcal{K}_I$ be applied to frequent trajectory pattern mining?
\end{enumerate}

To answer the first two questions, in addition to comparing the proposed kernels $\mathcal{K}_I$ with $\mathcal{K}_G$, they are also compared with 
\begin{itemize}
    \item Three commonly used distance measures: fastDTW \cite{fastDTW}, Hausdorff distance, and Frech\`{e}t distance; 
    \item Deep representation learning t2vec \cite{t2vec-ICDE2018}.
\end{itemize}

These measures and representations are examined mainly in the context of anomaly detection.
We examine the effectiveness of the above measures and representations by using them in four existing point anomaly detectors: IDK, GDK, LOF, and OCSVM, as already described in Table \ref{tab:variants} and Section \ref{sec:kad}.  Each of them assumes the role of point anomaly detector $\mathcal{F}$ in Algorithm \ref{alg:IDK2}. 


In addition, three deep learning anomaly detectors: Deep anomaly detectors GM-VSAE \cite{GM-VSAE-ICDE2020}, EncDec-AD \cite{EncDec-AD}, and Anomaly Transformer\footnote{Since trajectory is similar to time series, we are trying to apply time series anomaly detector on trajectory to see whether it can work well.} \cite{AnomalyTransformer} are also included in the comparison.


In the task of anomalous sub-trajectory detection, we compare Algorithm \ref{alg:subtraj-detector2} with a well-cited work TRAOD \cite{Partition-Detect-ICDE2008}.

For frequent sub-trajectory pattern mining, RegMiner \cite{RegMiner} is used to compare with Algorithm \ref{alg:fpm} since it is the most recent work and performs better than other previous methods.

\textbf{Evaluation metrics}.
ROC-AUC score is used to evaluate the detection accuracy of the anomaly detectors.  

\textbf{Parameter settings}.
The search ranges for all parameters in the experiments are given in Table \ref{search ranges}. 

\begin{table}[htbp]
    \centering
    \tabcolsep=2mm
    \begin{tabular}{cccc}
    \toprule
      & Parameter search ranges\\\midrule
     $d_W$,$d_H$,$d_F$ & No parameter tuning is required\\
    $\mathcal{K}_I$  & $\psi \in \{2^q|q=1,2,\dots,10\}$; $t=100$\\
      $\mathcal{K}_G$  & Nystr\"{o}m setting: n\_components $=100$;\\
      & $\sigma \in\{2^q|q=-10,-9,\dots,5\}$ \\
    t2vec &cellsize $\in\{25,50,100\}$; minfreq $\in\{10,50,100\}$;\\
    &hiddensize $\in\{2^q|q=6,7,8,9,10\}$ \\
    & hiddensize is the number of features used\\ 
    
    \midrule
    LOF  &$k \in \{1,\lfloor0.1n\rfloor,\lfloor0.2n\rfloor,\dots,\lfloor 0.9n \rfloor\}$; \\
    &$n$ is the number of trajectories\\
       \multicolumn{2}{l}{GDK,SVM \hspace{2cm} $\sigma\in\{2^q|q=-10,-9,\dots,5\}$}\\ 
        & other default settings are used in SVM\\
       IDK  & $\psi \in \{2^q|q=1,2,\dots,10\}$; $t=100$\\
       
     \midrule
    GM-VSAE  & $C \in \{1, 5, 10, 20, 50, 80\}$; \\
    & $C$ is the number of Gaussian components \\
    
    \midrule
    EncDec-AD & $ c \in \{4, 40, 64, 128\}$ \\
    & LSTM layers $ \in \{1, 2, 4\}$ \\
    
    \midrule
    Anomaly & $ d_{model} \in \{128, 256, 512\}$ \\
    Transformer & $d_{model}$ is the channel number of hidden states \\

    \midrule
    Algorithm \ref{alg:subtraj-detector2} & $\psi = 4096 $; $t = 100$; $\tau  = 0$ on \textit{Flyingfox} \\ & $\psi = 2048 $; $t = 100$; $\tau  = 0$ on \textit{Curlews} \\
    TRAOD & $\varepsilon = 1$; $\theta = 0.1$  on \textit{Flyingfox}\\ & $\varepsilon = 1$; $\theta = 0.05$  on \textit{Curlews}\\ 
        & $\varepsilon$ is the threshold; $\theta$ is penalty coefficient\\ 
    \midrule
    Algorithm \ref{alg:fpm} & $\psi=1024, t=100, \gamma = 0.06$ on \textit{Cross};\\
    & $\psi=16, t=100, \gamma = 3$ on \textit{Casia} \\
    RegMiner &  $\sigma = 950$ on \textit{Cross}; \\
             & $\sigma = 15$ on \textit{Casia}; \\
                & $\sigma$ is the support threshold \\
    \bottomrule
    \end{tabular}
    \caption{Parameter search ranges}
    \label{search ranges}
\end{table}

The implementation of Isolation kernel described in \cite{ting2020-IDK} are used to produce $\mathcal{K}_I$.

The machine used in the experiments has two AMD7742 64-core CPUs \& 1024GB memory; and two GPUs RTX3090 24GB. The GPUs are used by t2vec \cite{t2vec-ICDE2018}, EncDec-AD\cite{EncDec-AD}, and GM-VSAE \cite{GM-VSAE-ICDE2020} only. 

\textbf{Datasets}.
We perform the evaluations on twelve datasets, among which four datasets (\textit{Cross} \cite{cross2009},
\textit{Traffic}\footnote{\url{https://min.sjtu.edu.cn/lwydemo/Trajectory\%20analysis.htm}} \cite{Tube-Droplet-PAMI2017},
\textit{Casia}\footnote{\url{https://github.com/mcximing/ACCV18\_Anomaly}} \cite{tDPMM-PMAI2013}, 
\textit{Detrac}\footnote{\url{https://detrac-db.rit.albany.edu/}}
\cite{CVIU_UA-DETRAC}) are used in previous anomaly detection works and two are classification datasets (\textit{Character}\footnote{\url{https://archive.ics.uci.edu/ml/datasets/Character+Trajectories}}, \textit{Vrut}\footnote{\url{https://www.th-ab.de/ueber-uns/organisation/labor/kooperative-automatisierte-verkehrssysteme/trajectory-dataset}}). 
The other six datasets (\textit{Baboons}, \textit{Curlews}, \textit{Wildebeest}, \textit{Vultures}, \textit{Flyingfox}, \textit{Sheepdogs}) are collected from MoveBank\footnote{\url{https://www.movebank.org/cms/movebank-main}}, where each dataset records trajectories of a kind of animal over a period. In MoveBank datasets, trajectories that deviate from the majority are manually labeled as anomalies (see the Appendix for details). The data characteristics of these datasets are given in Table \ref{tab:datasets_inf}.
\begin{table}[htbp]
    \centering
    \tabcolsep=2mm
    \begin{tabular}{lrrrrr}
    \hline
         Dataset & \#Points & min -- max $|X|$ & \#Traj & \#AT & \%AT  \\
    \hline
         Baboons & 1,020,107 & 30 -- 603 & 2,310 & 110 & 5\% \\
         Curlews &801,489 &488 -- 71,821& 42 & 9& 21\%\\
         Character & 446,643 & 109 -- 205 & 2,643 & 28 & 1\% \\
         Detrac &445,052 &11 -- 2,120 & 5,356&71&1\%\\
         Vrut & 407,402 &55 -- 1,257 & 1,168 &100 & 9\%\\
         Wildebeest& 279,082&138 -- 5,632 & 92&14&15\% \\
         Vultures & 212,485 & 172 -- 7,721 & 67 & 15 & 22\% \\
         Cross & 153,010 & 4 -- 30 & 11600 & 200 & 2\%\\
         Casia &143,383&16 -- 612& 1,500 & 24&2\%\\
         Flyingfox &132,252 &517 -- 4,768 & 62 & 11 &18\% \\
         Sheepdogs & 65,574 & 11 -- 3,501 & 538 & 23 & 4\% \\
         Traffic & 11,500& 50 -- 50 & 230 & 30 & 13\% \\
         
    \hline
    \end{tabular}
    \caption{Real-world datasets. AT: anomalous trajectories. min -- max $|X|$: the minimum and maximum numbers of points of individual trajectories in a dataset.}
    \label{tab:datasets_inf}
\end{table}

\begin{table*}[htbp]
    \centering
    \resizebox{\linewidth}{!}{
    \begin{tabular}{l|ccc|cc|cc|cc|cccc|c|c|c}
    \hline
     & \multicolumn{3}{c|}{Euclidean distance-based} & \multicolumn{6}{c|}{Distributional kernel $\mathcal{K}$ using mapping $\Phi$} & \multicolumn{4}{c|}{Deep rep. learning (t2vec)} & GM & EncDec & Anomaly\\ 
     \cline{2-14} 
    Dataset  	&	 LOF$_{W}$ 	&	 LOF$_{H}$	&	 LOF$_{F}$ 	&	 LOF$_{I}$ 	&	 LOF$_G$ 	&	 SVM$_{I}$ 	&	 SVM$_G$ 	&	 GDK$_G$ 	&	 IDK$_I$ & LOF & SVM & GDK & IDK & -VSAE& -AD & Transformer	 \\ \midrule
         Baboons 	&	.91	&	.90	& \textbf{.99}	&	.96	&	\textbf{.99}	&	.86	&	.71	&	.95	&  \textbf{.99} & .72 & .78 & .81 &	.80 & .60 &.84 & .84\\
         Curlews 	&	.67	 & .79	 & \textbf{.90}	&	.82	&	.72	&	.73	&	.78	&	.67	& .83	& .62 & .49 & .38 & .36 & .51 & .53 & \tiny OOM \\
         Character & .85 & .70 & .78 & .76 & .85 & .49 & .47 & .47 & \textbf{.88} & .76 & .82 & .83 & .82 & .55 & .81 & .60\\
         Detrac & .82 & .78 & .70 & \textbf{.84} & .56 & .83 & .54 & .63 & \textbf{.84} & .75 & .68 & .64 & .72 & .50 & .70 & \tiny OOM \\
         Vrut 	&	.91	& \textbf{.93}	& .87 &	.91	&	.91	&	.80	&	.85	&	.84	& .89 & .77 & .72 & .73 &  .77 & .50 & .87 & .59  \\
         Wildebeest 	&	.78	&	.78	&	 \textbf{.89}	&	.84	&	.72	&	.61	&	.61	&	.77	&	.82 & .75 & .75 & .73 & .77 & .53 & .59 & \tiny OOM\\	
         Vultures   	&	.79	&	.74	&	.83	&	.84	&	.85	&	.74	& .81 &	.81	&	\textbf{.87}	 & .76 & .71 & .75 & .78 & .51 & .69 & \tiny OOM\\
         Cross & .82 & .89 & .84 & .83 & .82 & .64 & .77 & .92 & \textbf{.94} & .62 & .83 & .87 & 83 & .50 & .50 & .55\\
         Casia  & .70 & .69 & .73 & .78 & .61 & .84 & .62 & .71 & \textbf{.89} & .64 & .65 & .66 & .69 & .50 & .82 & .51\\
         Flyingfox  	&	.85	&	.88	& \textbf{.96}	 	&	.84	&	.72	&	.80	&	.67	&	.74	&	.89	 & .83 & .70 & .71 & .67 & .52 & .76 & \tiny OOM\\
         Sheepdogs 	&	.81	&	.72	&	.56	&	 \textbf{.99} 	&	.93	&	.95	&	.71	&	.70	&	.98	& .98 & .98 & \textbf{.99} & \textbf{.99} & .50 & .83 & \tiny OOM\\
         Traffic & .75  & .62 & .81 & \textbf{.98} & .91 & .52 & .69 & .86 & .96  &  .77 & .63 & .71 & .76 & .52 & .96 & .91 \\
         
    \hline
        Average Rank: & 6.3 & 6.9 & 5.2 & 3.8 & 7.2 & 8.9 & 11.2 & 8.0 & 2.3 & 9.2 & 10.3 & 9.1 & 8.6 & 14.4 & 8.9 & -\\
    \hline
    \end{tabular}
    }
    \caption{ROC-AUC results of different methods. The anomaly detectors that employ kernel mean maps $\Phi$ derived from $\mathcal{K}_I$ \& $\mathcal{K}_G$ are denoted with subscripts $_I$ \& $_G$, respectively. SVM denotes  OCSVM \cite{OCSVM2001}. 
    Boldface indicates the best ROC-AUC score in each dataset. OOM denotes that an algorithm has `out of memory' error during execution. The last row shows the rankings of all detectors in each dataset, averaged over all datasets.}
    \label{tab:exp_res}
\end{table*}

\section{Experimental Results}
\label{sec_experiments}
We present the results of three applications: anomalous trajectory detection, anomalous sub-trajectory detection and frequent trajectory pattern mining in the following three subsections.

\subsection{Anomalous trajectory detection results}
\label{sec:anomaly-detection}
Table \ref{tab:exp_res} presents the detection accuracy results. 
The following are the main findings of the experiments.
\renewcommand{\theenumi}{\alph{enumi}}
\begin{enumerate}
    \item In terms of similarity measures: all three detectors GDK, LOF \& SVM employing $\mathcal{K}_I$ are ranked higher than those using $\mathcal{K}_G$ (see the `Distributional Kernel $\mathcal{K}$' column). LOF with $\mathcal{K}_I$ is also ranked better than that employing $d_W$, $d_H$ and $d_F$. 
    Despite having extra learning, t2vec is not competitive compared with all other measures on most datasets.
    
    \item In terms of anomaly detectors: IDK$_I$ (which employs $\mathcal{K}_I$) performs better than all other detectors. The closest contender is LOF$_I$ which also employs $\mathcal{K}_I$. Deep learning anomaly detectors,  GM-VSAE, EncDec-AD and Anomaly Transformer perform worse than LOF$_I$ and IDK$_I$. These deep learning results on anomalous trajectory detection are consistent with those on time series anomaly detection \cite{TimeseriesEvaluation-VLDB2022,TimeseriesEvaluation-VLDB2022-2}. Additional analyses are provided in the Appendix.

\end{enumerate}


Figure \ref{fig:significant-test} shows the result of the Friedman-Nemenyi test \cite{NemenyiTest-2006}, comparing IDK$_I$ with GDK$_G$, LOF$_F$ (the top-ranked LOF that uses a distance measure), LOF$_I$ (the top-ranked LOF that uses $\mathcal{K}$), t2vec+IDK (the top-ranked t2vec) and EncDec-AD (the top-ranked end-to-end deep anomaly detector). Although IDK$_I$ is not significantly better than LOF$_I$ and LOF$_F$, it is the only detector that is significantly better than EncDec-AD, GDK$_G$ and t2vec+IDK. 

\begin{figure}[thbp]
\centering
   \includegraphics[width=0.8\linewidth]{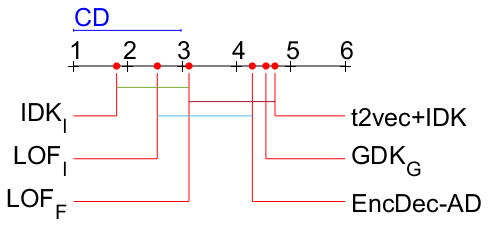}
   \caption{Friedman-Nemenyi test at 0.10 significance level. No significant difference if two detectors are connected by a CD line.}
   \label{fig:significant-test}
\end{figure}




\subsubsection{Time complexity}
For a dataset $D$ with $n$ trajectories and a total of $N$ points, the time complexity for IDK mapping is $\mathcal{O}(N\psi t)$ and for anomaly detection is $\mathcal{O}(n \psi t)$, where 
both $\psi$ and $t$ are parameters of IDK. See \cite{ting2020-IDK,ting2020-IDK-GroupAnomalyDetection} for more details.

\subsubsection{Scaleup test}
We perform a scaleup test using $10^2$ trajectories and $10^4$ trajectories randomly selected from the \textit{Cross} dataset, and the result is presented in Table \ref{tab:scaleup}. Key observations are:

\begin{enumerate}
    \item A striking difference between the three distance measures (the first three rows in the prep runtime ratio column) and the distributional kernels (the next two rows) that employ the same LOF:  each of the three distance measures runs three orders of magnitude slower than either $\mathcal{K}_I$ or $\mathcal{K}_G$.
    
    \item IDK$_I$ \& GDK$_G$ have linear runtime; LOF and SVM (using either $\mathcal{K}_I$ or  $\mathcal{K}_G$) have superlinear or quadratic runtime (see the AD runtime ratio column). 
    
    \item Because of using GPUs, t2vec, GM-VSAE and EncDec-AD have the lowest scaleup ratios; all other methods run on CPUs only. 
\end{enumerate}


A neural metric learning method has been proposed to speed up distance measures such as $d_W$ and $d_H$  by using an RNN to approximate distance measures: achieving 50x-1000x speedup at the cost of (degraded) 80\% accuracy of a distance measure \cite{MetricLearning-ICDE2019}. This method does not change our conclusion here on anomalous trajectory detection because using it weakens the accuracy of LOF for $d_W$, $d_H$ and $d_F$ we have obtained in Table \ref{tab:exp_res}.

\begin{table}[h]
    \centering
    \tabcolsep=2pt
    \begin{tabular}{l|rr|rr|rr}
    \hline
        & \multicolumn{2}{c|}{$10^2$ traj}    & \multicolumn{2}{c|}{$10^4$ traj} & \multicolumn{2}{c}{runtime ratio}\\\cline{2-7}
        & prep & AD & prep & AD& prep & AD\\
        \hline
        LOF$_W$ & 2 & .006 & 1081782 & 19 &	540891	&	3167 \\
        LOF$_H$ & 2 & .004 & 549055 & 14 &	274528	&	3500\\
        LOF$_F$ & 3 & .003 & 424061  & 12 &	141354	&	4000 \\
        LOF$_I$ & .9&.004 & 798&11&	887	&	2750\\
        LOF$_G$ & 1.8&.002 & 1663&14&	924	&	7000\\
        SVM$_I$ & 1.7&.001 & 1663&10&	978	&	10000\\
        SVM$_G$ & 2.1&.003 & 1663&10&	792	&	3333\\
        \hline
        IDK$_I$ & \multicolumn{2}{c|}{2.7} &  \multicolumn{2}{c|}{203} & \multicolumn{2}{c}{75} \\
        GDK$_G$ & \multicolumn{2}{c|}{1.2} & \multicolumn{2}{c|}{388} & \multicolumn{2}{c}{323}\\
        \hline
        t2vec+LOF & 281&.006 & 452&36 &	2	&	6000\\
        t2vec+SVM & 281&.024 & 452&93 &	2	&	3875\\
        t2vec+IDK& 281&.070 & 452&6.4 &	2	&	91\\
        t2vec+GDK & 281&.064 & 452&14.8 &	2	&	231\\
        \hline
        GM-VSAE & \multicolumn{2}{c|}{12} &  \multicolumn{2}{c|}{31} & \multicolumn{2}{c}{3} \\
        EncDec-AD & \multicolumn{2}{c|}{702} &  \multicolumn{2}{c|}{3908} & \multicolumn{2}{c}{6} \\
        
        Anomaly Transformer & \multicolumn{2}{c|}{105} & \multicolumn{2}{c|}{27542} & \multicolumn{2}{c}{262} \\
        \hline
    \end{tabular}
    \caption{Runtime (all in CPU secs except t2vec, GM-VSAE and EncDec-AD employing GPU). The preprocessing (prep) includes all calculations to produce a similarity matrix. AD is the runtime of the anomaly detector only. The ratio is the runtime for $10^4$ trajectories over that for $10^2$ trajectories. 
    }
    \label{tab:scaleup}
   \vspace{-2mm}
\end{table}

\subsection{Anomalous sub-trajectory detection} 
\label{sec:sub-detect2}
To evaluate the detection accuracy of Algorithm \ref{alg:subtraj-detector2}, we conduct an experiment on the \textit{Flyingfox} and \textit{Curlews} datasets. 

The ground-truth anomalous sub-trajectories in a dataset are identified using the following method objectively. A point in an anomalous trajectory is labeled anomalous if there is no normal trajectory in its  local neighborhood, and then contiguous anomalous points are linked into anomalous sub-trajectories (short sub-trajectories are discarded). 

We employ a commonly used Jaccard index \cite{jaccard1912distribution} to measure the matching between a detected anomalous sub-trajectory and a ground truth. It measures how well the two matched, the larger the index the better. 

All results on \textit{Flyingfox} are shown in Table \ref{tab:sub-detect-res-2}. Algorithm \ref{alg:subtraj-detector2} performs better than a highly cited sub-trajectory detection method TRAOD \cite{Partition-Detect-ICDE2008} in terms of Jaccard index; and it runs two orders of magnitude faster.

\begin{table}[htbp]
\label{tab:alg3_exp_res}
\centering
\begin{tabular}{c|cc|cc}
\hline
    & \multicolumn{2}{c|}{Jaccard index} & \multicolumn{2}{c}{Time(s)}  \\ 
    \hline
    & Alg2          & TRAOD & Alg2 & TRAOD \\ \hline
1   & 0.94          & 0.57  & 3.43 & 800                             \\
21  & 0.91          & 0.82  & 2.06 & 471                             \\
24  & 0.62          & 0.52  & 4.54 & 1130                           \\
26  & 0.85          & 0.80  & 2.36 & 764                             \\
34  & 0.90          & 0.81  & 1.78 & 581                            \\
38  & 0.71          & 0.60  & 4.38 & 1125                            \\
41  & 0.75          & 0.67  & 2.01 & 647                             \\
45  & 0.91          & 0.81  & 1.54 & 396                             \\
46  & 0.95          & 0.86  & 1.81 & 671                           \\
47  & 0.87          & 0.68  & 0.71 & 208                            \\
48  & 0.98          & 0.91  & 0.59 & 167                            \\ \hline
avg & 0.85         & 0.73  & 2.29 &  632                          \\ \hline
\end{tabular}
\caption{Algorithm $2$ versus TRAOD.}
\label{tab:sub-detect-res-2}
\end{table}

Table \ref{fig:subtraj-detection} shows the example results of anomalous sub-trajectory detection on the \textit{Flyingfox} and \textit{Curlews} datasets.

The example on the \textit{Flyingfox} dataset has two anomalous trajectories $Q_1$ \& $Q_2$. Algorithm \ref{alg:subtraj-detector2} using $\mathcal{K}_I$  identifies that $Q_1$ has two anomalous sub-trajectories (drawn in red; 
 and normal sub-trajectories are drawn in green); and $Q_2$ has one anomalous sub-trajectory.
 
 A well-cited method TRAOD \cite{Partition-Detect-ICDE2008} included parts of the normal sub-trajectories as anomalous sub-trajectories in both $Q_1$ \& $Q_2$; and only one anomalous sub-trajectory was detected in $Q_1$. 

On the larger \textit{Curlews} dataset consisting of more than 800,000 points, TRAOD took 2 days to identify the anomalous sub-trajectories of a given anomalous trajectory; whereas Algorithm \ref{alg:subtraj-detector2} using $\mathcal{K}_I$  completed it in 
less than 15 minutes. Besides, Algorithm  \ref{alg:subtraj-detector2} produces a more accurate result consistent with the ground truth as shown in the last row of Table \ref{fig:subtraj-detection}.

 \begin{table}[t]
 \tabcolsep=1pt
 \centering
  \begin{tabular}{c|cc}
    \hline
 &  Flyingfox   & Curlews\\
      \hline
    \begin{turn}{90} \quad \quad  Ground truth   \end{turn}  
&     \includegraphics[width=.47\linewidth]{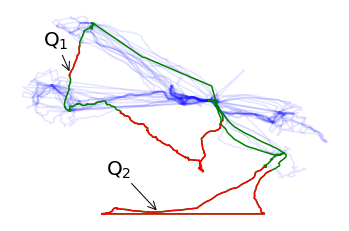} &
      \includegraphics[width=.47\linewidth]{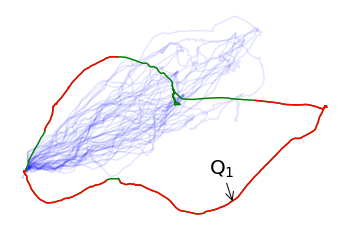}\\ \hline
  \begin{turn}{90} \quad \quad  TRAOD   \end{turn} 
 &  \includegraphics[width=.47\linewidth]{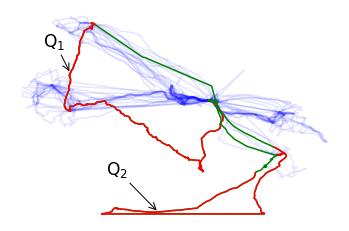}   &  \includegraphics[width=.47\linewidth]{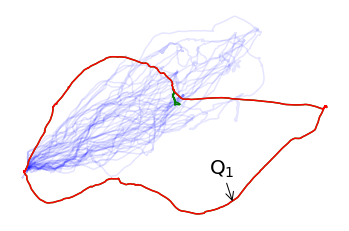}\\ \hline

\begin{turn}{90} \quad \quad  Algorithm \ref{alg:subtraj-detector2}   \end{turn} 
 &  \includegraphics[width=.47\linewidth]{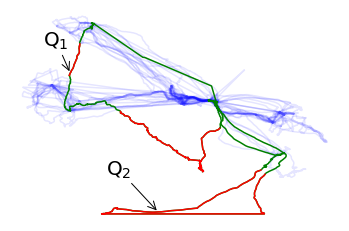}   &  \includegraphics[width=.47\linewidth]{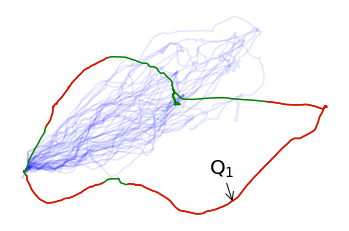}\\

\hline
\end{tabular}
    \caption{Anomalous sub-trajectory detection results of different algorithms on \textit{Flyingfox} and \textit{Curlews}. $Q_1$ \& $Q_2$ are two separate anomalous trajectories. In each anomalous trajectory, the detected anomalous sub-trajectories are colored in red; and normal sub-trajectories are in green. Every normal trajectory is colored in blue.}
    \label{fig:subtraj-detection}
\end{table}

In the absence of a powerful kernel like $\mathcal{K}_I$, TRAOD 
\cite{Partition-Detect-ICDE2008}, which uses a partition-and-detect framework, is a sensible method. As we have shown in Table~\ref{fig:subtraj-detection}, the sub-trajectories, as a result of the partitioning before the detection of anomalous sub-trajectories, are a coarse approximation. It is unable to produce the fine-grained sub-trajectories discovered by the proposed Algorithm \ref{alg:subtraj-detector2}. 
On the other hand, TRAOD \cite{Partition-Detect-ICDE2008} has high time complexity because it employs a combination of three distances (to represent the horizontal, vertical and angular distances) in the partition component to subdivide each trajectory into sub-trajectories. After the partitioning process, it employs LOF \cite{LOF-2000} with Hausdorff distance to identify anomalous sub-trajectories among all sub-trajectories. Trajectories with identified anomalous sub-trajectories are reported to be anomalous. TRAOD is a computationally expensive process because all computations are point-based (not distribution-based).

\subsection{Frequent sub-trajectory pattern mining}
\label{sec_exp_fpm}
Here we conduct an experiment to evaluate Algorithm~\ref{alg:fpm}.
We use the \textit{Cross} dataset which consists of nineteen clusters, and the \textit{Casia} dataset which consists of fifteen clusters. We also compared Algorithm \ref{alg:fpm} with an existing trajectory pattern mining method RegMiner \cite{RegMiner}. 

\begin{table}[htbp]
\renewcommand{\arraystretch}{1.5}
\begin{tabular}{l|ll}
\hline
      & \hspace{0.5cm}Algorithm \ref{alg:fpm} & \hspace{2cm}RegMiner    \\ \hline 
 \begin{turn}{90} \quad \quad Cross \end{turn} & \multicolumn{2}{l}{\includegraphics[width=.8\linewidth]{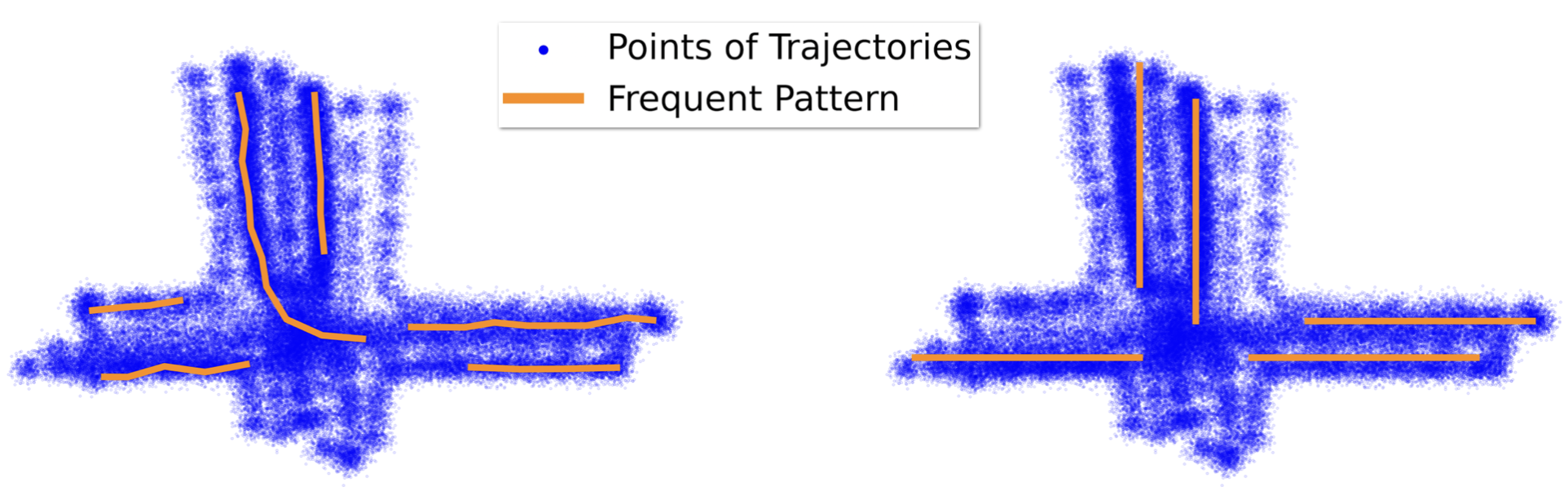}} \\ 
\hline
\begin{turn}{90} \quad \quad Casia \end{turn} & \multicolumn{2}{l}{\includegraphics[width=.8\linewidth]{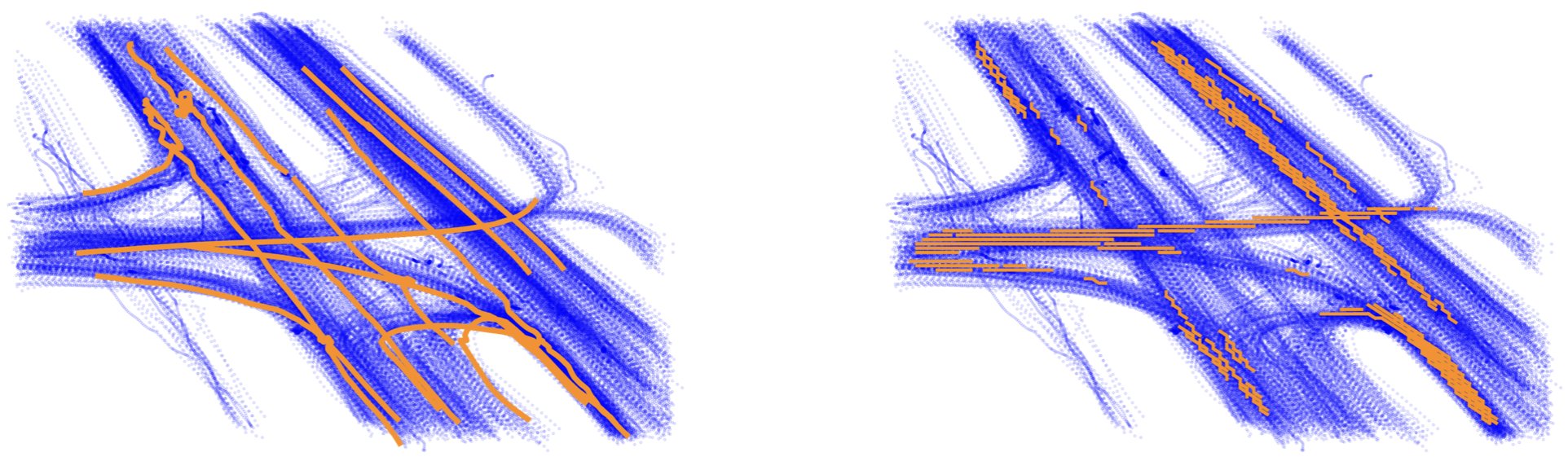}}
\end{tabular}
\caption{The first row shows the frequent sub-trajectory pattern mining results in \textit{Cross}, and the second row shows the results in \textit{Casia}. The left and right columns show the results of Algorithm \ref{alg:fpm} and RegMiner, respectively.}
    \label{tab:fpm_res}
\end{table}

\begin{table}[htbp]
\begin{tabular}{c|l|rrr}
\hline
Dataset                & Methods  & \#FP & Min length & Max length \\ \hline
\multirow{2}{*}{\textit{Cross}} & RegMiner & 23   & 112  &  224   \\
                       & Algorithm \ref{alg:fpm}     &  6    &   75  & 304    \\ \hline
\multirow{2}{*}{\textit{Casia}} & RegMiner &   643   &  9   &  28   \\
                       & Algorithm \ref{alg:fpm}     &   13   &  87   & 338 \\ \hline  
\end{tabular}
\caption{Frequent sub-trajectory pattern mining results on \textit{Cross} and \textit{Caisa}. \#FP denotes the number of frequent patterns each algorithm has detected. Min \& Max lengths denote the minimum and maximum lengths of the detected patterns, where the length is the sum of the piece-wise Euclidean distance of two adjacent points of a sub-trajectory.}
\label{tab:fpm_res_2}
\end{table}

Table \ref{tab:fpm_res} shows the visualized results of frequent sub-trajectory pattern mining. All points in a dataset are drawn in blue, which shows the locations those trajectories occupied. The more trajectories passed through, the darker the color. The discovered frequent patterns are drawn in orange.

The detailed results shown in Table \ref{tab:fpm_res_2} provide some interesting outcomes. RegMiner discovers significantly more frequent patterns than Algorithm \ref{alg:fpm}. But RegMiner's patterns are significantly shorter, especially in the \textit{Casia} dataset where the trajectories are sampled with significantly more points than those in \textit{Cross} (see the details in Table \ref{tab:datasets_inf}).
RegMiner's patterns have a maximum length of 28 only. In contrast, Algorithm 3 produces a maximum length of 338; even its shortest pattern with a length of 87 is three times longer than RegMiner's longest pattern.

This result is not surprising since RegMiner, like the PrefixSpan algorithm \cite{PrefixSpan} on which it is based in frequent pattern mining, produces many more short patterns than long patterns.  

\begin{figure}[htbp]
    \centering
    \includegraphics[width=.80\linewidth]{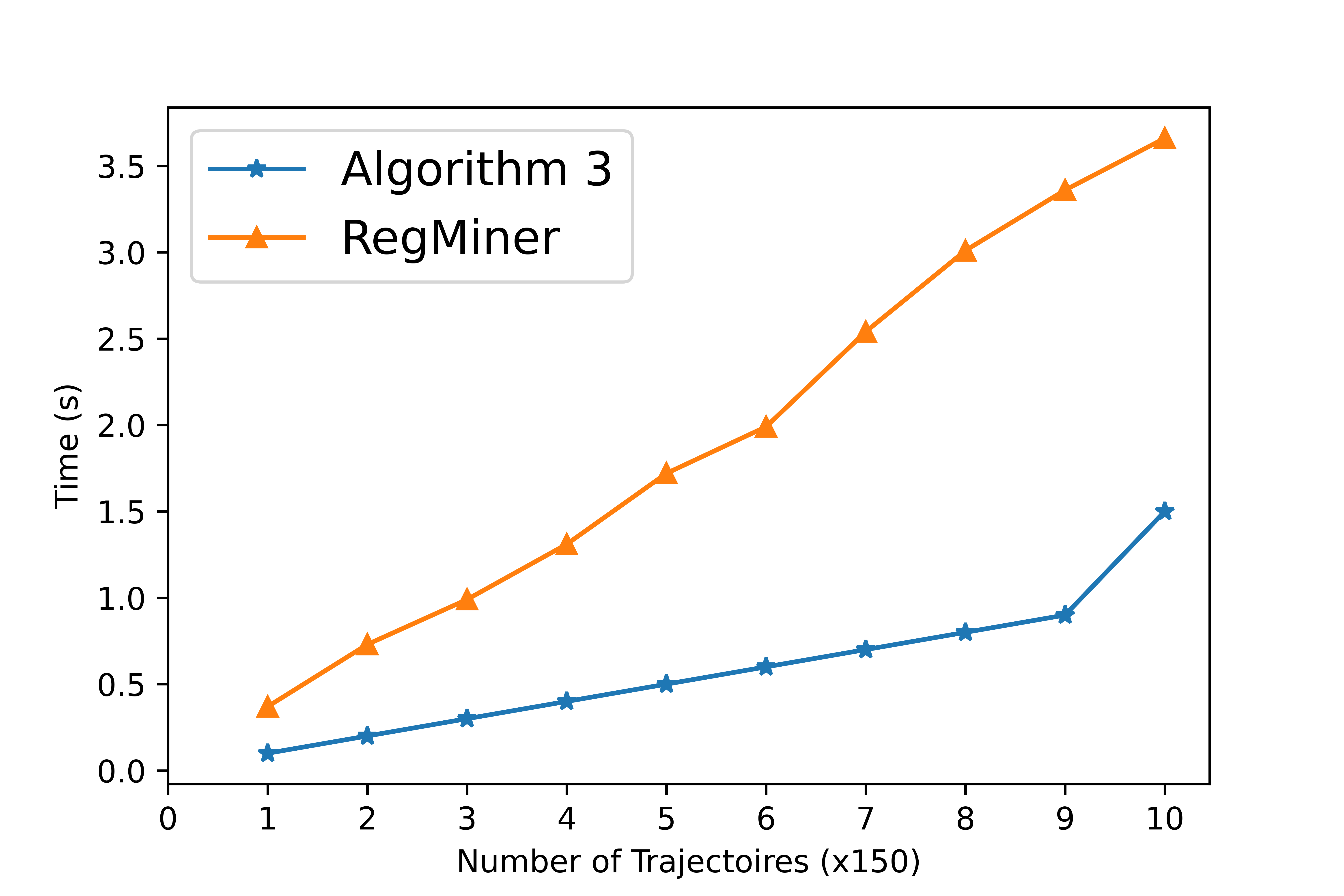}
    \caption{Runtime of Algorithm \ref{alg:fpm} and RegMiner on Casia}
    \label{fig:scaleup_test}
\end{figure}

 
\textbf{Scaleup test}. We conduct a scaleup test on \textit{Casia} for both Algorithm \ref{alg:fpm} and RegMiner. The result in Figure \ref{fig:scaleup_test} shows that both Algorithm \ref{alg:fpm} and RegMiner have linear time complexity, while the former runs faster than the latter.

\subsection{Discussion}
\label{sec_discussion}


It is interesting to note that deep learning methods t2vec \cite{t2vec-ICDE2018}, EncDec-AD\cite{EncDec-AD}, and Anomaly Transformer\cite{AnomalyTransformer} are not competitive with the proposed distributional kernel $\mathcal{K}_I$ without learning. This suggests that a powerful kernel such as $\mathcal{K}_I$ is more effective and efficient than deep learning methods for anomalous trajectory detection. This is mainly due to the use of distributional information and the data dependent property in $\mathcal{K}_I$.

The above result also raises three questions: (i) Can deep learning produce a representation or measure which is as powerful as $\mathcal{K}_I$? (ii) Can deep learning representations or measures have a data-dependent property like the one provided by $\mathcal{K}_I$? (iii) Can deep learning produce a measure $dist(\cdot,\cdot)$ which guarantees the uniqueness property: $dist(X,Y) = 0$ if and only if  $X=Y$. These are interesting topics for future research in deep learning.


Many existing measures (e.g., the set-based Hausdorff and Fr\`{e}chet distances) are based on iid implicitly. 
In contrast, our approach brings the iid assumption to the forefront and uses a distributional measure. The fact that it works well in practice indicates that iid  is a veritable assumption for valid trajectories in the real world. 

The proposed use of a distribution kernel for trajectories has been shown to work well in real-world datasets in Section \ref{sec_experiments}. The distributional kernel can fail in  some circumstances. 
Here we examine two circumstances in which a minor change can fix the problems. First, the distributional kernel does not distinguish between the trajectory of one cycle and another of multiple cycles on the same path.
If this difference is important, the length of each trajectory shall be added as an additional attribute. Second, if differentiating disparate traveling agents is important, then an additional agent attribute shall be included. These are minor tweaks to accommodate special needs, and they are not fundamental limitations of the proposed distributional  kernel.

\section{Conclusions}
\label{sec_conclusions}
The use of distributional kernels
is a paradigm shift in trajectory mining. Existing works are mostly point-based and do not consider distribution as a unit for computations. 

In addition to addressing the two fundamental weaknesses of existing measures for trajectories (mentioned in the Introduction section), we show (i) the power of the distributional kernel and its impacts in three applications; and (ii) the significance of the data-dependent property to lift the detection capability in datasets with clusters of varied densities. The data-dependent $\mathcal{K}_I$ is almost always better than the data-independent $\mathcal{K}_G$, even though both have the same uniqueness property. This shows the importance of the data-dependent property. 

$\mathcal{K}_I$ produces better detection accuracy than all other 
measures and representations mentioned in this paper. Coupled with the existing detector IDK, IDK$_I$ performs significantly better than four deep learning anomaly detectors. This is because only $\mathcal{K}_I$ has the uniqueness and the data-dependent properties. None of the deep learning anomaly detectors have been shown to have these properties. In addition, $\mathcal{K}_I$ runs orders of magnitude faster than the three existing distance measures. 

We also show that the proposed $\mathcal{K}_I$ based algorithms for anomalous sub-trajectory detection and frequent sub-trajectory pattern mining are simpler, faster, and more effective than the widely cited partition-and-detect method TRAOD and sequence pattern mining based method RegMiner.

\section*{Appendix}

Six datasets, i.e., Baboons, \textit{Curlews}, \textit{Wildebeest}, \textit{Vultures}, \textit{Flyingfox} and \textit{Sheepdogs} are collected from www.movebank.org/cms/movebank-main, which record trajectories of different animals over a time period. 
Trajectories in each dataset are extracted as follows: 

\textbf{Baboons}: Each original trajectory is a GPS-recorded activity of a baboon over half a month in August 2012. 
    Because the original has a very high sampling rate, we reduce the sampling rate by a factor of 1000.
    A small number of trajectories that deviate from the majority are considered as anomalous. 

\textbf{Curlews}: Each trajectory is an annual migration path of a curlew. A few trajectories which have different starting points or are not back to the same starting point are considered to be anomalous.
    
\textbf{Wildebeest}: Each trajectory is an annual migration path of a wildebeest; and a few trajectories have different routes are considered to be anomalous.

\textbf{Vultures}: Each trajectory is an annual migration path of a vulture. Those having no return trips or are too short are considered as anomalous.
    
\textbf{Flyingfox}: Each trajectory is a daily activity path of a flying fox. Those, which are significantly different from the majority or do not return to the starting point, are considered to be anomalous.
    
\textbf{Sheepdogs}: Trajectories are extracted between a long pause of a recording device. 515 trajectories belong to sheepdogs are normal trajectories; while 23 trajectories belong to a sheep are anomalies. An example visualization is shown in Figure~\ref{fig:sheepdogs}.

\section*{Additional analyses of deep learning}
This section provides the details of additional analyses on two  deep learning anomaly detectors \textbf{GM-VSAE} and \textbf{EncDec-AD}, and representation deep learning \textbf{t2vec}, focusing on the issue of the kind of datasets used for training.

\textbf{GM-VSAE} achieves ROC-AUC score at 0.69 on \textit{Baboons}, which is the worst among all methods listed in Table \ref{tab:exp_res}. Its ROC-AUC score on all other datasets are worse than that on \textit{Baboons}.
    GM-VSAE has been given the advantage of  using a training set of normal trajectories only because it aims to model normality \cite{GM-VSAE-ICDE2020}. All other methods in Table \ref{tab:exp_res} are trained using the given dataset that contains anomalous trajectories.
    
    GM-VSAE was reported to have high PR-AUC on two datasets only \cite{GM-VSAE-ICDE2020}. However, the result is an outcome of wrongly assigning normal trajectories as positive examples in computing the precision-recall curve (see their code at \url{https://git.io/JelML}, retrieved on 26 October 2021). This means that GM-VSAE is good at ranking many normal trajectories at the top. But this says nothing about its ability to detect anomalous trajectories. 
    
    Recall that GM-VSAE has one major drawback, i.e., its grid-based representation could not guarantee to have the uniqueness property such that two different  trajectories can potentially be mapped into the same series of tokens. We think that this is the main cause of its poor detection performance. 
    
\textbf{EncDec-AD}: We conducted an additional supervised version experiment for EncDec-AD by training the model with normal trajectories only. The experiment result shows that although the ROC-AUC score increases on some datasets, EncDecAD is still not competitive with IDK$_I$.
    
    

\textbf{t2vec}: We also conducted an experiment to train t2vec on the given dataset versus the set of normal trajectories only. The difference is small, and there is no suggestion that the latter will produce a better result. Their best results are still significantly worse than all results of Euclidean distance-based measures and distributional kernels (except one) shown in Table~\ref{tab:exp_res}.  


An interesting phenomenon is that IDK performs better when trained with the given dataset than that with normal trajectories only, unlike other detectors. The robustness of IDK to noise in the training set has been previously revealed \cite{ting2020-IDK}.


Our results on deep learning are consistent with those on time series anomaly detection \cite{TimeseriesEvaluation-VLDB2022,TimeseriesEvaluation-VLDB2022-2}, summarized below:

“.. deep learning approaches are not (yet) competitive despite their higher processing effort on training data.” \cite{TimeseriesEvaluation-VLDB2022}

“.. CNN and LSTM … are the third and the second-worst for sequence-based anomalies.” \cite{TimeseriesEvaluation-VLDB2022-2}

\newpage
\bibliographystyle{ACM-Reference-Format}
\bibliography{references}

\end{document}